%% file: icml2025.tex
\documentclass{article}
\usepackage[table]{xcolor}
\usepackage{microtype}
\usepackage{graphicx}
\usepackage{subfigure}
\usepackage{booktabs} %
\usepackage{tikz}
\usepackage{pgfplots}
\usepackage{multirow}
\usepackage{markdown}
\usepackage{verbatim}
\usepackage{listings}
\usepackage{float}
\usepackage{soul}
\usepackage{longtable}
\usepackage{supertabular,booktabs}
\usepackage{pifont}
\usepackage[most]{tcolorbox}
\usepackage[draft]{minted}
\usepackage{enumitem}

\usepackage{wrapfig}
\usepackage{hyperref}

\definecolor{lightblue}{RGB}{220,235,250}

\usepackage[accepted]{icml2025}

\usepackage{amsmath}
\usepackage{amssymb}
\usepackage{mathtools}
\usepackage{amsthm}
\usepackage{lipsum}
\usepackage{fontawesome}
\usepackage{marvosym}
\usepackage{xspace}
\usepackage{caption}

\usepackage[capitalize,noabbrev]{cleveref}

\theoremstyle{plain}

\theoremstyle{definition}

\theoremstyle{remark}

\usepackage[textsize=tiny]{todonotes}

\newcommand{\benchmark}{{\fontfamily{ppl}\selectfont MedXpertQA}\xspace}
\newcommand{\benchmarktext}{{\fontfamily{ppl}\selectfont MedXpertQA \textcolor{cyan!40!black}{Text}}\xspace}
\newcommand{\benchmarkmm}{{\fontfamily{ppl}\selectfont MedXpertQA \textcolor{cyan!40!red}{MM}}\xspace}
\newcommand{\benchtext}{{\fontfamily{ppl}\selectfont \textcolor{cyan!40!black}{Text}}\xspace}
\newcommand{\benchmm}{{\fontfamily{ppl}\selectfont \textcolor{cyan!40!red}{MM}}\xspace}

\newcommand{\bluehl}[1]{\sethlcolor{lightblue}\hl{#1}}
\newcommand{\orangehl}[1]{\sethlcolor{orange!10}\hl{#1}}
\newcommand{\yellowhl}[1]{\sethlcolor{yellow!9}\hl{#1}}

\newcommand{\tabincell}[2]{\begin{tabular}{@{}#1@{}}#2\end{tabular}}

\icmltitlerunning{MedXpertQA: Benchmarking Expert-Level Medical Reasoning and Understanding}

\begin{document}

\twocolumn[
\icmltitle{\benchmark: Benchmarking Expert-Level Medical Reasoning and Understanding}

\icmlsetsymbol{equal}{*}
\icmlsetsymbol{cor}{\Letter}

\begin{icmlauthorlist}
\icmlauthor{Yuxin Zuo}{equal,thu}
\icmlauthor{Shang Qu}{equal,thu,ailab}
\icmlauthor{Yifei Li}{thu}
\icmlauthor{Zhangren Chen}{thu}
\icmlauthor{Xuekai Zhu}{thu}
\icmlauthor{Ermo Hua}{thu}
\icmlauthor{Kaiyan Zhang}{thu}
\icmlauthor{}{}
\icmlauthor{Ning Ding}{cor,thu,ailab}
\icmlauthor{Bowen Zhou}{cor,thu,ailab}
\end{icmlauthorlist}

\icmlaffiliation{thu}{Tsinghua University, Beijing, China}
\icmlaffiliation{ailab}{Shanghai Artificial Intelligence Laboratory, Shanghai, China}

\icmlcorrespondingauthor{Ning Ding}{dn97@mail.tsinghua.edu.cn}
\icmlcorrespondingauthor{Bowen Zhou}{zhoubowen@mail.tsinghua.edu.cn}

\icmlkeywords{Machine Learning, ICML}

\begin{center}
    \faGlobe~{\fontfamily{ppl}\selectfont \textbf{Leaderboard:}}~\url{https://medxpertqa.github.io}\\
    \vskip 0.04in
    \faGithub~{\fontfamily{ppl}\selectfont \textbf{Code:}}~\url{https://github.com/TsinghuaC3I/MedXpertQA}
\end{center}
\vskip 0.15in

]

\printAffiliationsAndNotice{\icmlEqualContribution} %

\begin{abstract}

We introduce \benchmark, a highly challenging and comprehensive benchmark to evaluate expert-level medical knowledge and advanced reasoning.
\benchmark includes $4,460$ questions spanning $17$ specialties and $11$ body systems. It includes two subsets, \benchtext for text evaluation and \benchmm for multimodal evaluation.
Notably, \benchmm introduces expert-level exam questions with diverse images and rich clinical information, including patient records and examination results, setting it apart from traditional medical multimodal benchmarks with simple QA pairs generated from image captions.
\benchmark applies rigorous filtering and augmentation to address the insufficient difficulty of existing benchmarks like MedQA, and incorporates specialty board questions to improve clinical relevance and comprehensiveness.
We perform data synthesis to mitigate data leakage risk and conduct multiple rounds of expert reviews to ensure accuracy and reliability.
We evaluate $18$ leading models on \benchmark.
Moreover, medicine is deeply connected to real-world decision-making, providing a rich and representative setting for assessing reasoning abilities beyond mathematics and code.
To this end, we develop a reasoning-oriented subset to facilitate the assessment of \textit{o1-like} models.

\end{abstract}

\begin{figure}
    \centering
    \includegraphics[width=\linewidth]{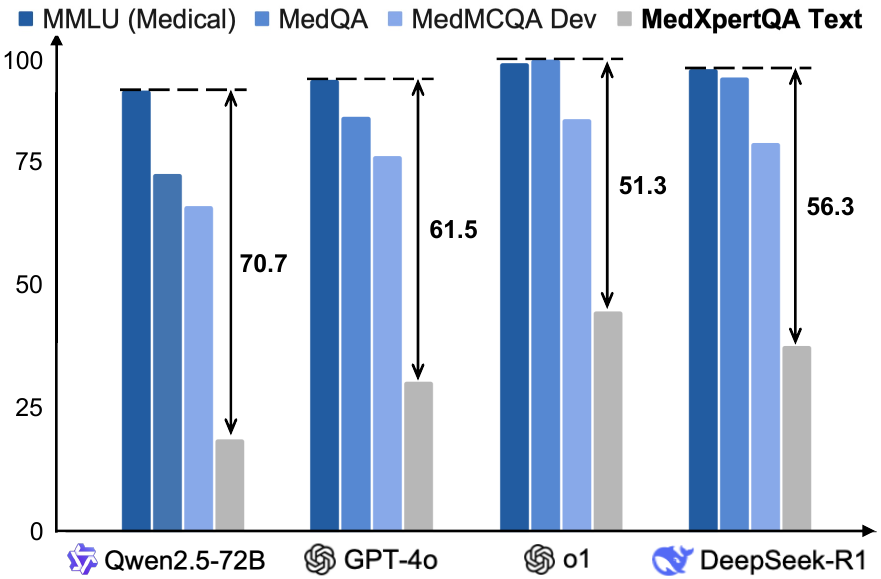}
    \caption{Performance of different models on \benchmarktext and other benchmarks.
    For other benchmarks, we report results for \texttt{o1-preview} in place of \texttt{o1}.
    Appendix~\ref{appx:full_performance_comparison} shows full results.
    }
    \label{fig:performance_comparison}
    \vspace{-4mm}
\end{figure}

\section{Introduction}

Large Multimodal Models (LMMs) have demonstrated promising potential in advancing general medical AI systems for applications in clinical scenarios~\cite{achiam2023gpt,liu2024visual,saab2024capabilities}.
However, current text and multimodal benchmarks for evaluating general medical AI capabilities have numerous limitations.

First, existing text medical benchmarks, such as PubMedQA~\cite{jin2019pubmedqa}, MedQA~\cite{jin2021disease}, MedMCQA~\cite{pal2022medmcqa}, and MMLU~(Medical)~\cite{wang2024mmlu}, lack comprehensive coverage of fine-grained and diverse real-world diagnostic scenarios, including highly specialized fields such as family and addiction medicine.
This lack of essential breadth limits the applicability of medical AI in thoroughly addressing realistic medical scenarios.
Moreover, these benchmarks fall short of sufficiently challenging current advanced AI, hindering progress toward reliable medical AI.
For instance, \texttt{o1-preview} has achieved $96\%$ and $99\%$ accuracy on MedQA-USMLE and MMLU Medical Genetics, respectively~\cite{nori2024medprompt}.

Second, traditional multimodal medical benchmarks, such as VQA-Rad~\cite{lau2018dataset}, Path-VQA~\cite{he2020pathvqa}, Slake~\cite{liu2021slake}, PMC-VQA~\cite{zhang2023pmc}, and OmniMedVQA~\cite{hu2024omnimedvqa}, are critically inconsistent with real-world clinical scenarios due to:
\textbf{1) Limited Scope and Insufficient Difficulty.}
These benchmarks solely evaluate basic visual perception and medical knowledge, neglecting the complexity of real-world medical tasks across different stages of the diagnosis process.
They fail to assess the expert-level knowledge and reasoning ability required for diagnostic decision-making and treatment planning.
\textbf{2) Lack of Authenticity and Clinical Relevance.}
Current benchmarks lack detailed clinical information and rely on automatically generated simple QAs paired with isolated medical images, diverging considerably from realistic clinical scenarios.
Medical exam questions used in existing text benchmarks present a promising solution, and \texttt{Med-Gemini}~\cite{saab2024capabilities} also demonstrates the significance of such evaluations.
However, the field still lacks such a systematic and high-quality benchmark.

\begin{figure*}[htbp]
    \centering
    \includegraphics[width=\linewidth]{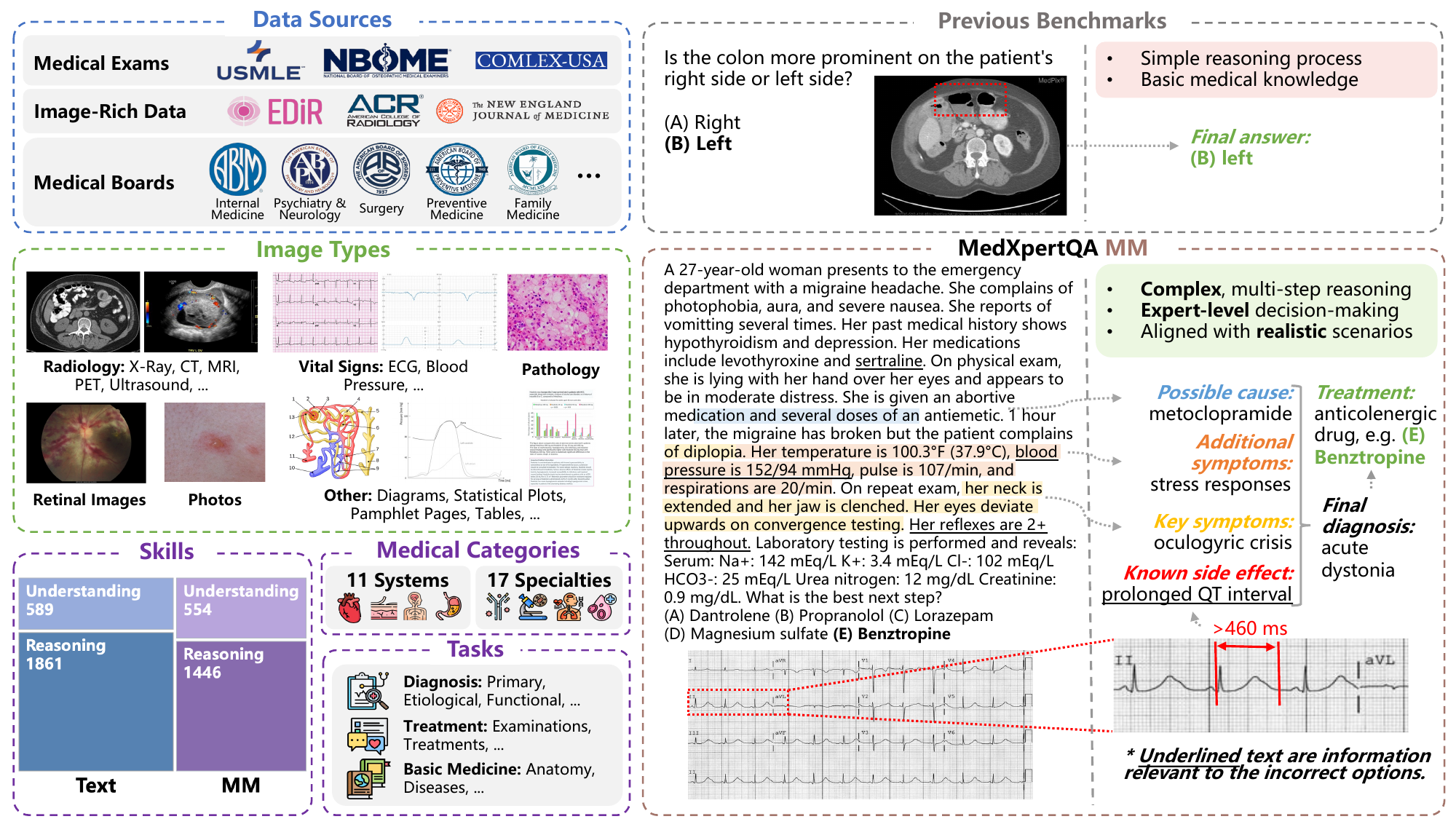}
    \caption{Overview of \benchmark.
    The left side illustrates its diverse data sources, image types, and question attributes.
    The right side compares typical examples from \benchmarkmm and a traditional benchmark~(VQA-Rad).
    }
    \label{fig:overview}
    \vspace{-4mm}
\end{figure*}

To address these challenges, we present \benchmark, a highly challenging and comprehensive medical multiple-choice benchmark.
It encompasses \benchmarktext for text-only evaluations and \benchmarkmm for multimodal assessments, making it suitable for a wide range of AI models.
Both subsets are currently the most challenging benchmarks in their respective fields.
Figure~\ref{fig:performance_comparison} presents model performance comparisons of \benchmarktext and other benchmarks.
\benchmarkmm includes diverse image types to simulate the wide range of visual information encountered in real-world diagnosis.
Figure~\ref{fig:overview} shows an overview.
Overall, \benchmark covers a wide range of medical specialties and systems and includes challenging real-world clinical tasks, enabling comprehensive evaluation of expert-level medical abilities.

\benchmark follows a rigorous and systematic construction process, including data collection, filtering, synthesis, and expert review.
\textit{We collaborate closely with medical practitioners throughout the design and review of \benchmark to ensure that its development reflects real-world medical practice and clinical reasoning.}
Specifically, we first curate a large-scale question bank from professional exams and textbooks, ensuring difficulty and diversity.
Sources include the United States Medical Licensing Examination~(USMLE) and the Comprehensive Osteopathic Medical Licensing Examination of the United States~(COMLEX-USA) for general medical evaluation, $17$ American specialty board exams for specialized scenarios, and $3$ image-rich sources, such as the NEJM Image Challenges.
Thus, it primarily aligns with medical practices and standards in the United States.
We subsequently perform extensive and multi-dimensional question filtering.
First, we conduct hierarchical filtering using an adaptive Brier score~\cite{zhu2024weighted} threshold based on thousands of human responses, calibrated to the difficulty ratings annotated by human experts.
We then filter questions based on $14$ sampling votes from $8$ AI experts for each question.
Additionally, we use probabilistic semantic similarity and exact matching precision to identify and remove highly similar questions, thereby enhancing robustness.
To ensure \benchmark is unseen for existing AI models, we use two proprietary models to rewrite questions and augment options, enhancing diversity and minimizing data leakage risk.
To increase difficulty and robustness, we also eliminate low-quality distractors (\textit{i.e.} easy and incorrect answer choices) and expand the number of options to $10$ and $5$ for \benchtext and \benchmm, respectively.
Finally, to mitigate potential errors such as factual inaccuracies, we engage medical experts with physician's licenses to review \benchmark, annotating and correcting any errors.

We evaluate $18$ proprietary and open-source LMMs and LLMs on \benchmark, including cutting-edge inference-time scaled models.
Results demonstrate that current leading models still exhibit limited performance on \benchmark, particularly in complex medical reasoning tasks.
In conclusion, our contributions are as follows:

\begin{itemize}[leftmargin=1em]
    \item We introduce \benchmark, a highly challenging and comprehensive medical multiple-choice benchmark.
    \benchmark integrates specialty-specific assessments into medical benchmarking and challenging exam questions with real-world clinical information into multimodal medical benchmarking.
    \item After rigorous filtering and data augmentation, \benchmark demonstrates exceptional difficulty and robustness.
    We further implement data synthesis to mitigate data leakage risk and conduct multiple rounds of expert reviews to ensure accuracy and validity.
    \item We evaluate $18$ leading models and analyze their limitations in medical capabilities.
    Moreover, we obtain a \textit{Reasoning} subset specifically designed for assessing medical reasoning abilities, which is well-suited for evaluating \textit{o1-like} reasoning models.
\end{itemize}

\section{Related Work}
\paragraph{Multimodal Medical Benchmarks}
Traditional multimodal medical benchmarks can be broadly categorized into two types: specialized and general purpose~\cite{chen2024gmai}.
Specialized benchmarks focus on a specific modality or medical domain.
VQA-Rad~\cite{lau2018dataset}, VQA-Med~\cite{ben2019vqa}, and Slake~\cite{liu2021slake} are primarily centered on radiology, while Path-VQA~\cite{he2020pathvqa} focuses on pathology.
These benchmarks provide extensive evaluation for their intended specialties, yet have a highly constrained scope and limited generalizability.
With advancements in LMMs, recent general-purpose benchmarks, such as OmniMedVQA~\cite{hu2024omnimedvqa} and GMAI-MMBench~\cite{chen2024gmai}, have provided more comprehensive evaluations.
However, these benchmarks cover limited medical image types, while real-world diagnostic scenarios encompass a wider variety of medical and even general images.
Furthermore, all traditional multimodal medical benchmarks mentioned above are constructed automatically based on image captions, deviating from real-world medical tasks.
MMMU (H \& M) Series~\cite{yue2024mmmu,yue2024mmmupro} alleviate this issue to some extent by introducing exam questions.
However, they are not specifically designed for the medical domain, and the difficulty is restricted to the university level.
Moreover, these benchmarks still lack detailed clinical information commonly encountered in real-world scenarios.

\textbf{Text Medical Benchmarks} \ \ 
Existing text medical benchmarks such as MedQA~\cite{jin2021disease}, MedMCQA~\cite{pal2022medmcqa}, MMLU~(Medical) Series~\cite{hendrycks2020measuring,wang2024mmlu} are derived from medical entrance and licensing exams, which primarily emphasize general medical knowledge and evaluation.
PubMedQA~\cite{jin2019pubmedqa} is a closed-domain benchmark, with questions generated from PubMed articles.
However, these benchmarks do not provide a thorough evaluation of highly specialized diagnostic scenarios to align sufficiently with real-world clinical practice.
Their difficulty has also become inadequate.

\begin{figure*}[t]
  \centering
  \includegraphics[width=\linewidth]{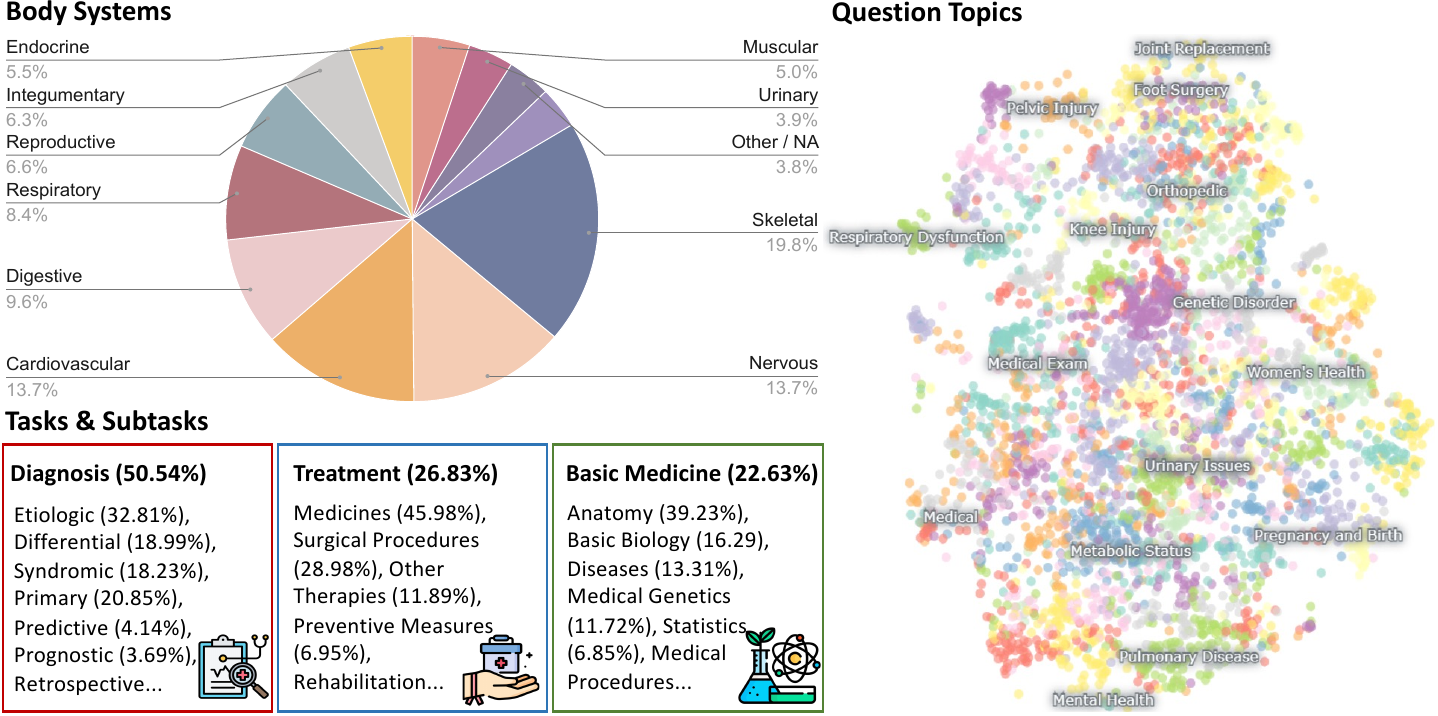}
  \caption{Attribute distributions of \benchmark showcase its diversity and comprehensiveness.}
    \label{fig:benchmark_distribution}
\end{figure*}

\section{\benchmark}
\subsection{Overview}
We introduce \benchmark, a universal medical benchmark consisting of challenging text and multimodal subsets \benchtext and \benchmm, which are each divided into a few-shot development set with $5$ questions and a test set.
Figure~\ref{fig:overview} shows an overview.
\benchmark is designed to assess expert-level medical knowledge and reasoning capabilities across $17$ medical specialties, $11$ human body systems, and $3$ task categories.
It includes a total of $4,460$ questions sourced from examinations at the medical licensing level or higher, of which $2,005$ are multimodal questions with a total of $2,839$ images.
It is the first to introduce medical specialty evaluations to better simulate realistic medical scenarios.
Moreover, \benchmarkmm is the first multimodal medical benchmark to incorporate challenging medical examination questions and real-world medical scenarios.
Meanwhile, due to medicine's rigorous requirements for highly reliable outcomes across a comprehensive scope, we refrain from solely prioritizing difficulty and instead aim for question diversity and range as well.
As a result, \benchmark both substantially challenges current models and showcases remarkable diversity across multiple dimensions.
Figure~\ref{fig:benchmark_distribution} illustrates its wide coverage.

\textbf{Medical Coverage}\ \ 
\benchmark collects questions from $17/25$ member board exams (specialties) of the American Board of Medical Specialties~\footnote{https://www.abms.org} to enable evaluation of highly specialized medical scenarios.
Moreover, applying the categorization in \citet{Liachovitzky2015differential}, we instruct an LLM to annotate each question with its most relevant human body system.
\benchmark covers all $11$ systems.

\textbf{Modal Diversity}\ \ 
The data in \benchmark covers a broad spectrum of modalities. Its questions and answer choices both incorporate structured data, such as tables, and semi-structured documents.
\benchmarkmm's images similarly demonstrate high diversity and wide coverage.
It not only encompasses medical imaging results obtained from various techniques in diverse formats, but also other image types such as diagrams, charts, and documents, fully covering the spectrum of visual information that human doctors are expected to analyze.
Specifically, \benchmark encompasses the following image categories: \textit{Radiology, Pathology, Medical Optical Imaging, Photos, Vital Signs, Diagrams, Documents, Charts, Tables, and Others}.

\textbf{Task Diversity}\ \ 
To enable in-depth analysis, we use \texttt{gpt-4o-2024-11-20} to annotate each question's \textit{core medical task} (Diagnosis, Treatment Planning, or Basic Medicine)~\cite{huang2025o1} and fine-grained subtask.
Appendix~\ref{appx:attribute-annotation-prompts} presents relevant prompts.

\subsection{Benchmark Construction}

\begin{table}[!t]
\centering
\resizebox{\linewidth}{!}{
\begin{tabular}{lccc}
\toprule
\textbf{Phase} & \benchmm & \benchtext & \textbf{Sum} \\
\midrule
Data Collection & $10,868$ & $26,675$ & $37,543$ \\
\midrule
\multicolumn{4}{c}{\textbf{Difficulty-Based Data Filtering}}\\
\midrule
Human Expert Filtering & $5,133$ & $5,013$ & $10,146$ \\
AI Expert Filtering & $2,125$ & $2,612$ & $4,737$ \\
\midrule
\multicolumn{4}{c}{\textbf{Quality-Based Filtering and Augmentation}}\\
\midrule
Edit Distance Filtering & $2,121$ & $2,592$ & $4,713$ \\
Semantic Similarity Filtering & $2,105$ & $2,578$ & $4,683$ \\
Question and Option Augmentation & $2,105$ & $2,578$ & $4,683$ \\
Multi-Rounds Expert Review \& Delete & $2,005$ & $2,455$ & $4,460$ \\
\bottomrule
\end{tabular}
}
\caption{Data statistics after different construction phases.}
\label{tab:data_construction}
\vspace{-4mm}
\end{table}

To construct \benchmark, we begin with data collection, followed by a carefully designed and resource-intensive procedure involving data filtering, synthesis, and expert review.
Table~\ref{tab:data_construction} presents dataset statistics across different construction phases.
After initial data collection and filtering based on question difficulty, the reductions in question count occur during the \textit{Similarity Filtering} and \textit{Expert Review} stages, which remove $54$ and $223$ questions, respectively.
While the early stages primarily optimize for difficulty, these later stages focus on improving data quality.
Given that \benchmark is constructed from authoritative and high-quality sources, filtering based on difficulty naturally has a more substantial impact on dataset size compared to quality-based filtering.
Additionally, during the expert review phase, reviewers directly corrected the errors in most of the flagged questions instead of deleting them.

\paragraph{Data Collection}
We begin by constructing a large-scale question bank, sourcing difficult multiple-choice questions from authoritative medical examinations and textbooks.
Previous work primarily relied on USMLE questions for training and evaluation~\cite{jin2021disease}.
We expand the scope by including questions from COMLEX, another major medical licensing examination in the U.S., to capture the unique challenges of medical image interpretation in orthopedic practice.
To further evaluate multimodal medical capabilities, we incorporate questions from the American College of Radiology (ACR) DXIT and TXIT exams, the European Board of Radiology (EDiR) exams, and the New England Journal of Medicine (NEJM) Image Challenge.
In addition, we collect questions from $17$ American medical specialty board exams, spanning a wide array of common medical specialties.
Ultimately, we collect $37,543$ questions.
We additionally obtain human response distributions and expert annotations, including explanations and difficulty ratings.

\paragraph{Data Filtering}
We conduct \textit{AI \& Human Expert Filtering} to identify questions that challenge both humans and AI.
Subsequent \textit{Similarity Filtering} further enhances robustness.

\textbf{Step 1: AI Expert Filtering.} \ \ 
We employ $8$ models, categorized as basic or advanced, as AI experts to vote on and filter questions.
First, each basic AI expert performs $4$ sampling attempts for each question. If any expert answers a question correctly in all attempts, the question is deemed too simple and removed.
Second, questions that are answered incorrectly by all AI experts are retained.
This method effectively differentiates between questions that leading AI models can solve and those that remain challenging.
Appendix~\ref{appx:ai_models} lists models used in this phase.

\textbf{Step 2: Human Expert Filtering.}\ \ 
We use prior and posterior human expert annotations to identify questions that challenge humans.
Using human response distributions, we first assess each question's posterior difficulty by calculating its Brier score, a widely applied metric of prediction accuracy~\cite{zhu2024weighted}.
We consider the question's response distribution over all answer choices as the answer prediction.
Given the prediction vector $\mathbf{\hat{y}}$ and label vector $\mathbf{y}$, the Brier score $B$ is:
\begin{align}
B = \frac{1}{N} \sum_{i=1}^{N} (y_i - \hat{y_i})^2,   
\end{align}
where $N$ is the number of options, $y_i$ is the label for option $i$ ($0$ or $1$), and $\hat{y_i}$ is the proportion of responses selecting option $i$.
A lower Brier score indicates a more accurate overall prediction distribution, suggesting the question is easier.
Compared to accuracy, the Brier score accounts for the response rates of all options, providing a more precise difficulty measurement.
Subsequently, we normalize the prior difficulty ratings annotated by medical experts and categorize questions into $3$ levels, each associated with an adaptive Brier score threshold for stratified sampling.
Higher-rated questions are assigned higher Brier score thresholds, with the maximum threshold set at the $25$th and the minimum set at the $3$rd percentile of all scores.
Approximately $16.78\%$ of questions lack the annotations above and thus do not undergo human expert filtering.

\textbf{Step 3: Similarity Filtering.}\ \ 
A key factor in achieving robust evaluation is ensuring high diversity and avoiding repetitive assessments.
Therefore, we filter data by identifying question pairs with extremely high semantic or edit distance similarities.
For semantic similarity, we use \texttt{MedCPT-Query-Encoder}~\cite{jin2023medcpt}, a medical embedding model developed from \texttt{PubMedBERT}~\cite{pubmedbert}, to obtain sentence representations of all questions, then compute pairwise cosine similarities.
Since similarities are one-dimensional and do not follow a Gaussian distribution, we use the Interquartile Range~(IQR) to identify outliers with abnormally high similarity.
For detected outlier question pairs, we compare their difficulty annotations and remove the easier question.
\textit{Similarity Filtering} plays a critical role in ensuring benchmark robustness.
Traditional visual medical benchmarks, such as Slake~\cite{liu2021slake}, often rely on rigid, template-based question generation, which limits question diversity.
This is problematic since model performance can vary greatly depending on question type, and repetitive questions allow models to exploit template-specific shortcut patterns, leading to unexpected performance gains.
In contrast, since \benchmark was not generated from fixed templates, it exhibits high question diversity from the start, resulting in relatively few questions being removed at this stage.
Nonetheless, even minimal filtering is indispensable, as it removes near-duplicate questions that could otherwise cause benchmark overfitting and distort evaluation results.

\begin{table*}[!t]
\centering
\resizebox{\textwidth}{!}{
\begin{tabular}{lcccccccc}
\toprule
\textbf{Benchmark}  &
 \textbf{\# Size} &\textbf{\# Images} &
\textbf{\# Rate} &
\tabincell{c}{\textbf{\# Avg} \\ \textbf{~Lens}} &
\tabincell{c}{\textbf{\# Image} \\ \textbf{~Types}} &
\textbf{Annotation} &
\tabincell{c}{\textbf{Clinical} \\ \textbf{Scenarios}} & \textbf{Specialties}\\ 
\midrule
\rowcolor{lightblue}VQA-Rad~\cite{lau2018dataset}         &  $451$    &$204$    &          $0.45$   &$14.61$ &$1$ & Automatic & $\times$  &$\times$  \\
\rowcolor{lightblue}VQA-Med~\cite{ben2019vqa}             &  $500$    &$500$    &          $1.00$   &$8.96$ &$1$ & Automatic & $\times$  &$\times$  \\
\rowcolor{lightblue}Path-VQA~\cite{he2020pathvqa}         &  $6,719$  &$858$  &          $0.13$   &$15.38$ &$1$ & Automatic & $\times$  &$\times$  \\
\rowcolor{lightblue}Slake-En~\cite{liu2021slake}          &  $1,061$  &$96$  &          $0.09$   &$13.97$ &$1$ & Automatic & $\times$  &$\times$  \\
\rowcolor{orange!10}PMC-VQA~\cite{zhang2023pmc}           &  $33,430$ &$29,021$ &          $0.87$   &$61.84$ &$2$ & Automatic & $\times$  &$\times$  \\
\rowcolor{orange!10}OmniMedVQA~\cite{hu2024omnimedvqa}    &  $127,995$&$118,010$&          $0.92$   &$42.40$ &$4$ & Automatic & $\times$  &$\times$  \\
\rowcolor{orange!10}GMAI-MMBench~\cite{chen2024gmai}      &  $21,281$  &$21,180$  &        $1.00$    &$49.85$ &$4$ & Automatic & $\times$  &\checkmark  \\
\rowcolor{yellow!9}MMMU (H \& M)~\cite{yue2024mmmu}      &  $1,752$  &$1,994$  &          $1.14$   &$83.56$ &$8$ & Expert    & $\times$  &$\times$  \\
\rowcolor{yellow!9}MMMU-Pro (H \& M)~\cite{yue2024mmmupro} &  $346$    &$431$    &          $1.25$   &$107.08$ &$7$ & Expert    & $\times$  &$\times$  \\
\midrule
\benchmarkmm                          &  $2,000$&$2,852$&   $1.43$&$149.35$&$10$& Expert    & \checkmark  &\checkmark  \\
\bottomrule
\end{tabular}
}
\caption{\looseness=-1 Comparisons with existing multimodal medical benchmarks.
\benchmarkmm demonstrates high complexity through its high average question length, a large number of images, and diverse image categories.
\bluehl{The first category} includes commonly used traditional benchmarks.
\orangehl{The second category} includes recently emerging and more comprehensive benchmarks.
\yellowhl{The third category} includes general multimodal benchmarks that include medicine-related subsets but are not specifically designed for medical fields.}
\label{tab:compare_multimodal_bench}
\end{table*}

\begin{table}[!ht]
\centering
\resizebox{\linewidth}{!}{
\begin{tabular}{lcccc}
\toprule
\textbf{Benchmark} &
\textbf{\# Size} &
\tabincell{c}{\textbf{\# Avg} \\ \textbf{~Lens}} &
\tabincell{c}{\textbf{Clinical} \\ \textbf{Scenarios}} & \textbf{Specialties} \\ \midrule
\rowcolor{lightblue}PubMedQA~\cite{jin2019pubmedqa}                       & $1,000$   & $328.41$ & $\times$  &$\times$  \\
\rowcolor{lightblue}MedMCQA-Dev~\cite{pal2022medmcqa}                  & $4,183$                   & $53.84$ & $\times$  &$\times$  \\
\rowcolor{lightblue}MedQA-USMLE~\cite{jin2021disease}                  & $1,273$                   & $215.46$ & \checkmark  &$\times$  \\
\rowcolor{yellow!9}MMLU~(Med.)~\cite{hendrycks2020measuring}          & $1,089$                   & $100.07$ & $\times$  &$\times$  \\ 
\rowcolor{yellow!9}MMLU-Pro~(Med.)~\cite{wang2024mmlu}          & $586$                   & $166.63$ & $\times$  & $\times$  \\ \midrule
\benchmarktext & $2,450$& $257.37$ & \checkmark  & \checkmark  \\ \bottomrule
\end{tabular}
}
\caption{Comparisons with text medical benchmarks.
}
\label{tab:compare_text_bench}
\vspace{-4mm}
\end{table}

\paragraph{Question and Option Augmentation}
Since questions in \benchmark are from publicly available sources, we perform question and option augmentation to mitigate possible data leakage issues.
Option augmentation also increases both the difficulty and robustness of \benchmark.
For question augmentation, we instruct the LLM to rephrase the question through alternative expressions or structural adjustments while preserving all original information.
We emphasize professional language style, factual accuracy, and objectiveness.
For option augmentation, we first identify and remove low-quality distractors with low human response rates while retaining the correct option and at least two distractors.
Most multimodal questions consist of $5$ options which are often image-dependent and therefore cannot be feasibly expanded.
Therefore, we set the total number of options for \benchmarktext and \benchmarkmm to $10$ and $5$, respectively.
The generated distractors should be reasonable in the question's context, challenge medical experts, and remain consistent with the original options in both language and content style.
We use \texttt{gpt-4o-2024-11-20} and \texttt{claude-3-5-sonnet-20241022} in this stage, and Appendix~\ref{appx:data-augmentation-prompts} shows all prompts.
Section~\ref{sec:data-leakage} compares the leakage risk before and after data augmentation to verify the effectiveness of this process.

\paragraph{Expert Review}
Finally, medical experts with physician's licenses review each question in \benchmark, assessing the general quality of the original and augmented questions and options based on comprehensive guidelines.
Objectives of the question review include identifying missing information, detecting factual inaccuracies, and removing extra content.
Options review primarily emphasizes evaluating the options' validity and context-appropriateness.
Appendix~\ref{appx:expert-review} provides a detailed description of this procedure.

\subsection{Medical Reasoning Benchmarking}
OpenAI's o1 models advance reasoning capabilities, which have been extensively validated in math and coding~\cite{wu2024comparative}.
However, evaluation in specialized domains like medical reasoning has been insufficient, as relevant benchmarks remain underdeveloped.
Recent works developing \textit{o1-like} medical reasoning models with reinforcement learning~\cite{chen2024huatuogpt} or inference-time scaling~\cite{huang2025o1} similarly face limitations.
Benchmarks chosen, such as MedQA, contain questions assessing medical knowledge only (see \textit{Example of MedQA (Understanding)} in Appendix~\ref{appx:question-cases}), hence are suboptimal for isolating the model's medical reasoning ability.
Therefore, while leading \textit{o1-like} models show promise in medicine~\cite{xie2024preliminary,nori2024medprompt},
systematic evaluations specifically focused on their reasoning-based medical proficiency remain lacking.

Moreover, medicine tightly interweaves professional knowledge with complex reasoning.
Clinical reasoning is multifactorial~\cite{yazdani2019five}, requiring synthesizing diverse information, navigating uncertainty~\cite{patel2024navigating}, and engaging in heterogeneous cognitive processes~\cite{shin2019reasoning}.
These characteristics are represented in \benchmark through complex, information-rich tasks such as multiple-choice style differential diagnosis~\cite{seller2011differential} (see Appendix~\ref{appx:question-cases} and ~\ref{appx:error-cases}).
Therefore, we underscore the importance of identifying reasoning-oriented tasks to facilitate fine-grained performance feedback and accurate assessment.
On \benchmark, we use \texttt{gpt-4o} to annotate questions according to whether they require complex reasoning.
As seen in Appendix~\ref{appx:attribute-annotation-prompts}, we instruct the LLM to categorize complicated, reasoning-heavy questions as \textit{Reasoning}.
In contrast, other questions that involve little to no reasoning and instead assess skills such as medical knowledge and image perception are categorized as \textit{Understanding}.
As shown in~Figure~\ref{fig:overview}, within both \benchtext and \benchmm, a majority of questions focus on \textit{Reasoning}.

We initially considered having human experts annotate \textit{Reasoning} and \textit{Understanding} labels.
However, we found that LLMs can perform this task reliably when provided with clear and precise prompt instructions.
Specifically, we provide expert-written answers and explanations collected from data sources and explicitly emphasize the distinction between reasoning complexity and general difficulty in our prompt design, as shown in Table~\ref{tab:reasoning-annotation-prompt}).
This extensive guidance enables a simplified form of annotation under expert supervision.
Following automatic labeling, we engage human experts to conduct a review.
For a randomly selected $10\%$ subset of the questions, $490$ from \benchtext and $400$ from \benchmm, the reviewers identified $28$ and $11$ questions, respectively, that were incorrectly labeled.
This corresponds to an error rate of approximately only $4.3\%$.

\subsection{Comparisons with Existing Benchmarks}
Tables~\ref{tab:compare_multimodal_bench} and~\ref{tab:compare_text_bench} compare two subsets of \benchmark with existing benchmarks.
Traditional multimodal benchmarks deviate from real-world clinical tasks, reflected in the limited number of image types, low image-to-question ratios, and automatically generated questions and annotations.
Meanwhile, MMMU (H \& M) Series primarily based on university-level subject exams, falls short in scope, difficulty, and specificity to the medical domain.

\looseness=-1 \benchmarkmm demonstrates advantages in length and image richness.
It is the first multimodal medical benchmark to incorporate real-world clinical scenarios, resulting in a substantial increase in question length.
Meanwhile, the rich image information in \benchmarkmm can be seen from its broadest range of image types, highest number of images within same-scale benchmarks, and highest image-to-question ratio.

\benchmarktext is the first text medical benchmark to purposefully account for medical specialty assessment, supporting evaluations of highly specialized medical scenarios.

\begin{table*}[!t]
\centering
\resizebox{.935\linewidth}{!}{
\begin{tabular}{@{}lccccccccc@{}}
\toprule
 \multirow{2}{*}{$\mathbf{Model}$} &\multicolumn{3}{c}{\textbf{\benchmarktext}}&
  \multicolumn{3}{c}{\textbf{\benchmarkmm}}&
  \multirow{2}{*}{$\mathbf{Avg}_{R}$} &
  \multirow{2}{*}{$\mathbf{Avg}_{U}$} &
  \multirow{2}{*}{$\mathbf{Avg}$}  \\   \cmidrule(lr){2-7}&$\mathbf{Reasoning}$ & $\mathbf{Understanding}$  & $\mathbf{Avg}$ & $\mathbf{Reasoning}$ & $\mathbf{Understanding}$  & $\mathbf{Avg}$ &       &       &        \\ \midrule
\rowcolor{lightblue}Expert~(Pre-Licensed)                &$41.74$         & $45.44$          & $42.60$     & $45.76$         & $44.97$          & $45.53$     & $43.48$     & $45.20$     & $43.92$      \\ \midrule
\multicolumn{10}{c}{Inference-Time Scaled Large Multimodal Models} \\ \midrule
 o1$^\ddag$                &$46.24$         & $39.66$          & $44.67$     & $52.78$         & $65.45$          & $56.28$     & $49.09$     & $52.21$     & $49.89$      \\
 QVQ-72B-Preview     &$22.08$ & $20.71$  & $21.76$ & $33.54$ & $33.57$  & $33.55$ & $27.09$ & $26.95$ & $27.06$ \\\midrule
\multicolumn{10}{c}{Vanilla Large Multimodal Models} \\ \midrule
 GPT-4o            &\cellcolor{blue!40}{$30.63$} & \cellcolor{blue!40}{$29.54$} & \cellcolor{blue!40}{$30.37$}& \cellcolor{blue!40}{$40.73$}& \cellcolor{blue!40}{$48.19$}  & \cellcolor{blue!40}{$42.80$}& \cellcolor{blue!40}{$35.05$}& \cellcolor{blue!40}{$38.58$}& \cellcolor{blue!40}{$35.96$} \\
 Claude-3.5-Sonnet &$19.88$& \cellcolor{blue!20}{$25.81$} & \cellcolor{blue!20}{$21.31$}& $33.33$& $32.85$  & $33.20$& $25.76$& $29.22$& $26.65$ \\
 Gemini-1.5-Pro    &$19.18$& $21.22$ & $19.67$& $32.85$& $37.36$ & $34.10$& $25.16$& $29.05$& $26.16$ \\
 GPT-4o-mini       &$17.09$& $20.20$ & $17.84$& $28.22$& $27.62$ & $28.05$& $21.95$& $23.80$& $22.43$ \\
 Gemini-2.0-Flash     &\cellcolor{blue!20}{$20.53$}& $20.71$ & $20.57$& \cellcolor{blue!20}{$35.48$}& \cellcolor{blue!20}{$41.70$} & \cellcolor{blue!20}{$37.20$}& \cellcolor{blue!20}{$27.06$}& \cellcolor{blue!20}{$30.88$}& \cellcolor{blue!20}{$28.04$} \\ 
 Qwen2.5-VL-72B      &$17.89$& $18.17$ & $17.96$& $29.53$& $31.05$ & $29.95$& $22.98$& $24.41$& $23.35$ \\
 Qwen2-VL-72B      &$16.39$& $18.68$ & $16.94$& $25.86$& $34.84$  & $28.35$& $20.53$& $26.51$ & $22.07$ \\
\bottomrule
\end{tabular}
}
\caption{Performance of LMMs on \benchmark.
$^\ddag$ indicates the result is evaluated on a sampled subset.
$\mathbf{Avg}_{R}$ and $\mathbf{Avg}_{U}$ indicate the average performance on the $\mathbf{Reasoning}$ and $\mathbf{Understanding}$ subsets, respectively.
}
\label{tab:main}
\end{table*}

\begin{table}[!ht]
\centering
\resizebox{.95\linewidth}{!}{
\begin{tabular}{@{}lccc@{}}
\toprule
$\mathbf{Model}$ & $\mathbf{Reasoning}$ & $\mathbf{Understanding}$ & \textit{$\mathbf{Avg}$} \\ \midrule
\rowcolor{lightblue}\textcolor{gray}{Expert~(Pre-Licensed)}      &  \textcolor{gray}{$41.74$} &  \textcolor{gray}{$45.44$} &  \textcolor{gray}{$42.60$} \\
\textcolor{gray}{o1$^\ddag$}      &  \textcolor{gray}{$46.24$} &  \textcolor{gray}{$39.66$} &  \textcolor{gray}{$44.67$} \\
\textcolor{gray}{QVQ-72B-Preview}      &  \textcolor{gray}{$22.08$} &  \textcolor{gray}{$20.71$} &  \textcolor{gray}{$21.76$} \\ \midrule
\multicolumn{4}{c}{Inference-Time Scaled LLMs}                                \\ \midrule
o3-mini$^\ddag$      &  $37.63$ &  $36.21$ &  $37.30$ \\
DeepSeek-R1      &  $37.88$ &  $37.35$ &  $37.76$ \\
QwQ-32B-Preview  &  $18.70$ &  $15.79$ &  $18.00$ \\
\midrule
\multicolumn{4}{c}{Vanilla LLMs}                                              \\ \midrule
DeepSeek-V3      &  \cellcolor{blue!40}{$23.91$}&  \cellcolor{blue!20}{$24.96$}&  \cellcolor{blue!20}{$24.16$}\\
Claude-3.5-Haiku &  $16.71$&  $21.05$&  $17.76$\\
LLaMA-3.3-70B    &  \cellcolor{blue!20}{$23.86$}&  \cellcolor{blue!40}{$26.49$}&  \cellcolor{blue!40}{$24.49$}\\
LLaMA-3-70B-UltraMedical &  $20.03$&  $21.05$&  $21.80$\\
Qwen2.5-72B      &  $18.54$&  $20.03$&  $18.90$\\
Qwen2.5-32B      &  $14.02$&  $18.34$&  $15.06$\\
\bottomrule
\end{tabular}
}
\caption{Performance of LLMs on \benchmarktext.}
\label{tab:main_text}
\end{table}

\begin{table}[!t]
\centering
\resizebox{.95\linewidth}{!}{
\begin{tabular}{@{}lccc@{}}
\toprule
\textbf{Phase}  & \textbf{Perplexity $\uparrow$} & \textbf{Rouge-L $\downarrow$} & \textbf{Edit Distance Similarity $\downarrow$} \\ \midrule
Before & $1.03E+218$  & $0.1893$  & $0.2691$                   \\
After  & $1.35E+247$  & $0.1664$  & $0.2416$                   \\ \bottomrule
\end{tabular}
}
\caption{Data leakage analysis results.}
\label{tab:ana_data_leakage}
\vspace{-2mm}
\end{table}

\section{Experiments}

\subsection{Implementation Details}

We evaluate all models using zero-shot CoT prompting~\cite{kojima2022large} unless otherwise specified.
For answer cleansing, we follow the script provided by~\citet{kojima2022large}.
Appendix~\ref{appx:evaluation-case} shows a case of evaluation.
We employ greedy decoding for output generation if available, ensuring result stability.
For reasoning models with specific evaluation requirements, we follow their respective instructions.
Appendix~\ref{appx:evaluation_settings} presents additional implementation details.
We could not evaluate \texttt{o1} and \texttt{o3-mini} on the full \benchmark due to costs.
Instead, for both \benchmarktext and \benchmarkmm, we sample $10$\% of questions from the \textit{Reasoning} and \textit{Understanding} subsets respectively.
The seed is set to $42$.

\subsection{Models}
We evaluate leading LMMs and LLMs on full \benchmark and \benchmarktext, respectively.
We include both proprietary and open-source models, and also evaluate advanced inference-time scaled models$^\dag$ with a focus on the latest progress in medical reasoning capabilities.

\textbf{Large Multimodal Models}: o1-2024-12-17$^\dag$~\cite{jaech2024openai}, QVQ-72B-Preview$^\dag$~\cite{qvq-72b-preview}, GPT-4o-2024-11-20~\cite{hurst2024gpt}, GPT-4o-mini-2024-07-18~\cite{hurst2024gpt}, Claude-3.5-Sonnet-20241022~\cite{TheC3}, Gemini-1.5-Pro-002~\cite{team2024gemini}, Gemini-2.0-Flash-Exp~\cite{gemini2flash}, Qwen2.5-VL-72B~\cite{qwen2.5-VL}, Qwen2-VL-72B~\cite{Qwen2VL}.

\textbf{Large Language Models}: o3-mini$^\dag$~\cite{gpt-o3-mini}, DeepSeek-R1$^\dag$~\cite{guo2025deepseek}, QwQ-32B-Preview$^\dag$~\cite{qwq-32b-preview}, DeepSeek-V3~\cite{liu2024deepseek}, Claude-3.5-Haiku-20241022~\cite{TheC3}, LLaMA-3.3-70B-Instruct~\cite{dubey2024llama}, LLaMA-3-70B-UltraMedical~\cite{zhang2024ultramedical}, Qwen2.5-72B-Instruct~\cite{yang2024qwen2}, Qwen2.5-32B-Instruct~\cite{yang2024qwen2}.

\subsection{Main Results}
Tables~\ref{tab:main} and~\ref{tab:main_text} show the main results.
Overall, the low accuracies of evaluated models demonstrate \benchmark's ability to pose challenges to state-of-the-art models.

\textbf{Human Performance Evaluation}\ \ 
We establish a baseline denoted as \texttt{Expert (Pre-Licensed)}, referring to upper-year medical students or trainees preparing for licensing examinations such as USMLE, as shown in Table~\ref{tab:main}.
This baseline is derived from aggregated response distributions based on answers provided by these examinees, collected from the original data sources.
Since \textit{Question and Option Augmentation} preserves all core clinical content, its impact on human accuracy is minimal.
Accordingly, performance on the final questions closely aligns with that on the original versions.
Most questions have substantial response counts, with a large portion of the dataset containing at least $500$ responses per question and some reaching up to $\mathbf{238}$k. Fewer than $200$ final questions lack response data.
These characteristics ensure that the resulting statistics provide stable and representative estimates of human performance.

\textbf{Comparison of LMMs}\ \ 
Among vanilla LMMs, \texttt{GPT-4o} consistently performs best across all subsets.
\texttt{Gemini-2.0-Flash} is the highest-scoring vanilla LMM after \texttt{GPT-4o} with an impressive performance on \benchmarkmm, highlighting its advantage in multimodal tasks.
We also observe that the highly representative open-source LMM, \texttt{Qwen2.5-VL-72B}, outperforms \texttt{GPT-4o-mini}, yet still falls behind \texttt{GPT-4o}.

\textbf{Comparison of LLMs}\ \ 
\texttt{DeepSeek-R1} shows the strongest performance among LLMs, most notably on the \textit{Reasoning} set, where it substantially outperforms other models.
The performance of \texttt{Qwen2.5-32B}, particularly on the \textit{Reasoning} set, is nearly equivalent to random selection, so we limit the evaluation to models of this scale and above.

\textbf{Medical Reasoning Performance} \ \  
Table~\ref{tab:main} indicates that the textual reasoning performance of vanilla LMMs other than \texttt{GPT-4o} is relatively close, while their visual reasoning capabilities show marked differences, suggesting that visual perception plays a crucial role in approaching \benchmark questions.
Moreover, vanilla LLMs and LMMs consistently perform worse on the \textit{Reasoning} subset than on the \textit{Understanding} subset.
However, this gap noticeably narrows for \textit{o1-like} inference-time scaled models.
\textit{This underscores the inherent challenges associated with medical reasoning, as well as the high quality of our \textit{Reasoning} subset.}
In other words, our annotations effectively produce a challenging reasoning-oriented subset that is well-suited for evaluating \textit{o1-like} reasoning models.
Additionally, while models generally score higher on \benchmarkmm, this can be attributed to the smaller number of answer choices for \benchmarkmm and more rigorous filtering parameters used for constructing \benchmarktext.
Therefore, direct performance comparisons between the two subsets are not informative.

\begin{figure*}[!t]
    \centering
    \input{inference_time_scaling}
    \caption{Performance of three groups of models on \benchmark.}
    \label{fig:inference}
    \vspace{-4mm}
\end{figure*}
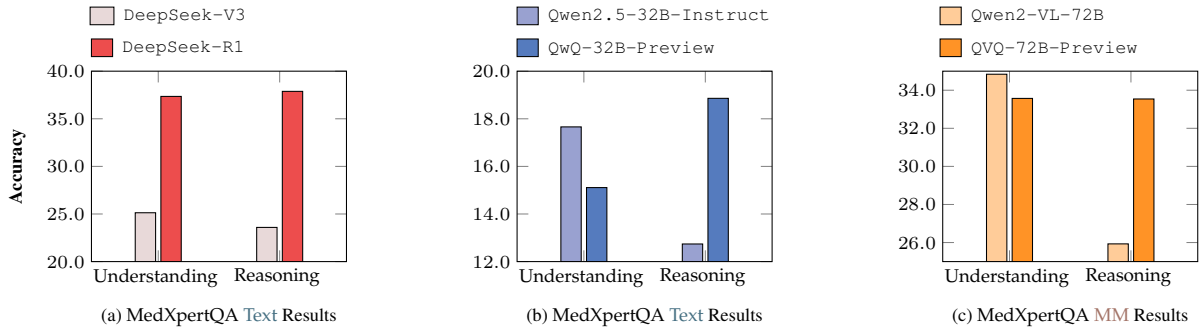

\section{Analysis}

\subsection{Data Leakage}
\label{sec:data-leakage}

To evaluate whether \textit{Question and Option Augmentation} reduces leakage risk, we follow~\citet{xu2024benchmarking} to use perplexity (PPL) and N-gram accuracy (ROUGE-L and edit distance similarity) as metrics.
We make certain adaptations and simplifications to assess leakage risk at the instance level before and after data synthesis.
After concatenating the original question with a specific prompt such as ``Answer:'' to form the input, we calculate the PPL of the model's output.
Moreover, to assess whether the evaluated model's rationale is similar to the corresponding explanations we collected, we compute the ROUGE-L and edit distance similarity between the output and the explanation.
We analyze the outputs of \texttt{GPT-4o}, since it is the most effective among all vanilla LMMs, possibly reflecting a higher risk of data leakage.

\begin{figure}[!t]
    \centering
    \includegraphics[width=\linewidth]{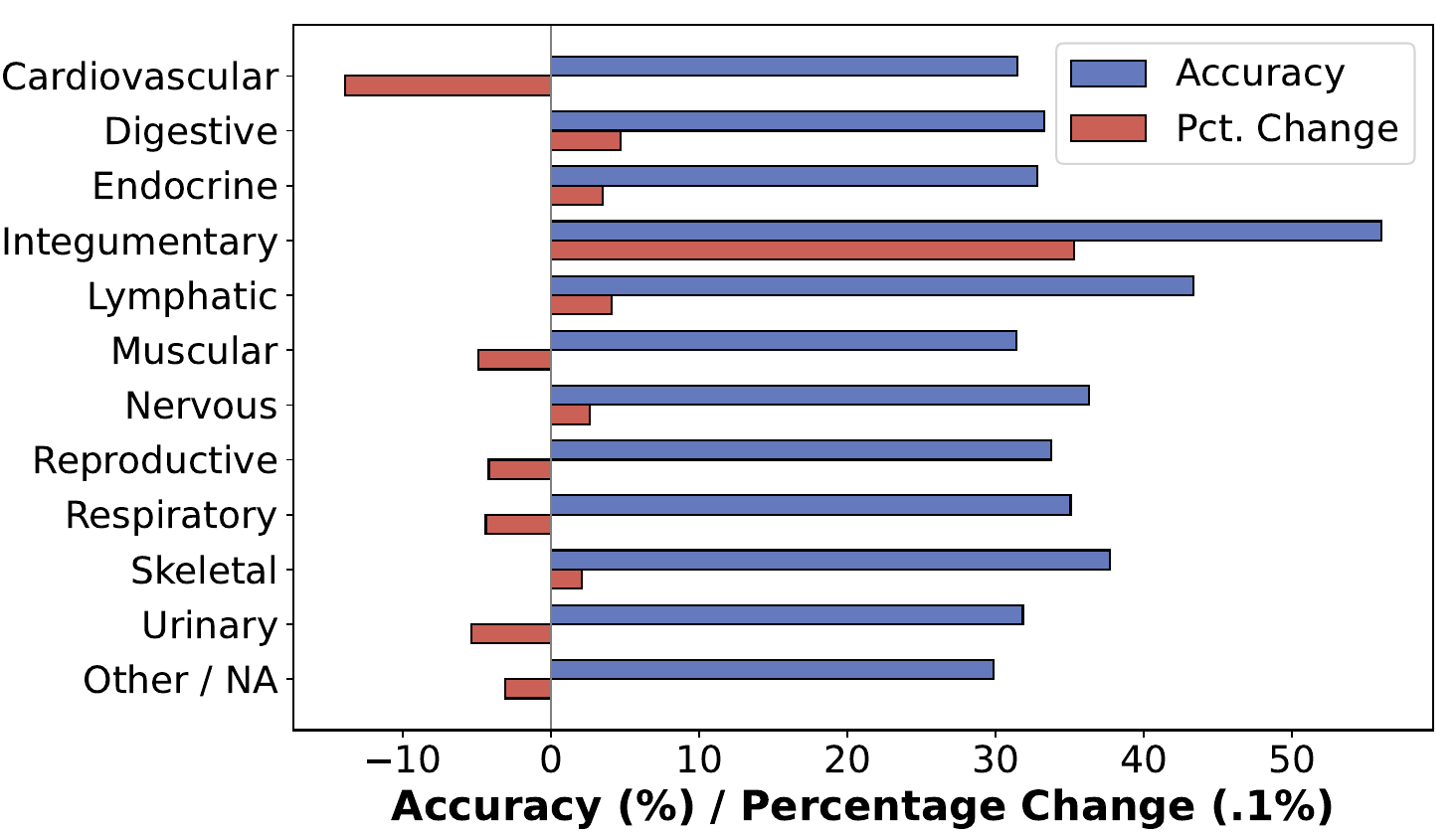}
    \caption{
    System analysis results.
    ``Accuracy'' bars show accuracies on each system's questions.
    ``Percentage Change'' bars show the gap between the percentages of each system's questions in the correctly answered set and the full set.
    }
    \label{fig:system}
    \vspace{-4mm}
\end{figure}

\looseness=-1 Table~\ref{tab:ana_data_leakage} shows results.
Based on the threshold set by~\citet{xu2024benchmarking}, the data leakage risk before synthesis is already relatively low, which can be attributed to two factors:
1) Questions in \benchmark are derived from difficult medical exams that may have not yet been considered for training;
2) Even questions already used for training may be difficult for the model to learn due to their complexity~\cite{lin2024rho}.
Leakage risk is further reduced following data synthesis, particularly indicated by a notable increase in perplexity.
\textit{Results demonstrate that \benchmark has a low risk of data leakage and can objectively assess model ability.}

\subsection{Impact of Inference-Time Scaling}

We employ two groups of Qwen-Series models for comparative analysis to investigate the impact of inference-time scaling on the models' performance on challenging medical tasks.
Specifically, they represent two text and multimodal \textit{o1-like} models, along with their corresponding backbones.
We similarly assess and compare \texttt{DeepSeek-R1} and \texttt{DeepSeek-V3}.

\vspace{-0.2cm}
\begin{itemize}[leftmargin=1em]
    \item \texttt{DeepSeek-R1} \textit{\textbf{vs}} \texttt{DeepSeek-V3}: Evaluation on \benchmarktext.
    \item \texttt{QwQ-32B-Preview} \textit{\textbf{vs}} \texttt{Qwen2.5-32B-Instruct}: Evaluation on \benchmarktext.
    \item \texttt{QVQ-72B-Preview} \textit{\textbf{vs}} \texttt{Qwen2-VL-72B}: Evaluation on \benchmarkmm.
\end{itemize}
\vspace{-0.2cm}

Figure~\ref{fig:inference} illustrates the performance of the three groups across \benchmarktext and \benchmarkmm.
The accuracy variation observed in each subplot is consistently more pronounced in the \textit{Reasoning} set.
Furthermore, accuracy on the \textit{Reasoning} set exhibits a clear upward trend, while performance on the \textit{Understanding} set fluctuates, with occasional declines.
This suggests that, even without additional medical training data, inference-time scaling can improve complex medical reasoning skills.

\subsection{Medical Insights}
We analyze \texttt{GPT-4o}'s performance based on the annotated system labels to derive fine-grained medical insights.
Figure~\ref{fig:system} shows that the proportion of \textit{Integumentary} questions in the correctly answered set markedly increases compared with the full question set. We also observe a much higher accuracy on these questions than questions on other systems.
This suggests that \texttt{GPT-4o} has a deeper understanding and stronger performance on \textit{Integumentary}-related questions.
In contrast, it exhibits lower accuracy on the \textit{Cardiovascular} subset, with a noticeable proportion decline in the correct set, indicating suboptimal capability in this regard.

\section{Conclusion}

\looseness=-1 We introduce \benchmark, a highly challenging and comprehensive medical multiple-choice benchmark evaluating expert-level knowledge and reasoning in real-world clinical scenarios.
\benchmark encompasses diverse medical specialties, body systems, and clinical tasks.
It addresses critical gaps in current benchmarks, including inadequate coverage of medical specialties, insufficient difficulty, and limited clinical relevance.
By incorporating expert-level medical examination questions rooted in comprehensive clinical information, \benchmarkmm marks a crucial advancement in multimodal medical benchmarking.
We mitigate data leakage risk through data synthesis and perform expert review to ensure accuracy and validity.
We benchmark $18$ leading models and analyze their system-specific medical capabilities.
Additionally, we construct a reasoning-oriented subset, which demonstrates that current AI systems still face challenges when tackling complex medical reasoning tasks.

\section*{Acknowledgements}

We thank anonymous reviewers for their insightful comments and suggestions.
This work is supported by National Science and Technology Major Project (2023ZD0121403)
, Young Elite Scientists Sponsorship Program by CAST (2023QNRC001), and 
National Natural Science Foundation of China (No. 62406165).

\section*{Impact Statement}

This paper introduces \benchmark, a comprehensive medical benchmark to advance the field of healthcare application-driven machine learning.

The deployment of AI in healthcare raises ethical concerns, encompassing issues such as data privacy, algorithmic biases, and the potential for excessive dependence on automated systems.
To mitigate these risks, it is crucial to prioritize transparency, explainability, and continuous validation by healthcare professionals. Furthermore, AI systems must undergo ongoing assessment to prevent the perpetuation of existing healthcare disparities.

While \benchmark demonstrates considerable promise for advancing medical AI, it also underscores the importance of responsible development and oversight to ensure that these technologies are applied ethically and equitably within healthcare settings.

\bibliography{icml_2025}
\bibliographystyle{icml2025}

\newpage
\appendix
\onecolumn
\section{Leakage Prevention Statement}
\begin{tcolorbox}[colback=lightblue!80,breakable,colframe=black]

\begin{itemize}[leftmargin=1em]
    \item \textcolor{red}{The data of \benchmark is strictly intended for model or human medical evaluation purposes only. \textbf{Please DO NOT use it for any form of training}, including training on synthesized or rephrased versions of the data.}

    \item To mitigate the potential data leakage risks, we refrain from releasing the data sources and request that you \textbf{do not share any example of \benchmark online}, whether in plain text, image, or any other format.

    \item We obtained all data from \textbf{freely and publicly accessible} sources. We only retained a small percentage of source data, and all questions were subject to question rephrasing and option shuffling, fully ensuring that \benchmark complies with \textbf{U.S. fair use laws}.
\end{itemize}
\end{tcolorbox}

\section{Discussions}

\subsection{Ethical Considerations}

To evaluate the reliability of \benchmark under real-world clinical variability, we consider the impact of input incompleteness on model performance.
Specifically, we propose future assessments using semantically equivalent clinical scenarios with systematically omitted salient information to quantify the sensitivity of model predictions to such variations.
This is motivated by the stochastic nature of LLMs and their reliance on prompt phrasing.
Furthermore, we highlight that benchmark performance does not imply clinical readiness.
\benchmark, while constructed through rigorous expert filtering and augmentation, remains a synthetic evaluation framework.
Safe use of medically oriented language models requires external safeguards beyond benchmark scores, including uncertainty estimation, structured oversight, and integration with clinical workflows. These considerations are essential for aligning benchmark utility with responsible use in healthcare contexts.

\subsection{Comparison with HLE~(Med)}

To illustrate the difficulty of \benchmark, we compare it against HLE~\cite{phan2025humanity}.
HLE is derived from original questions contributed by nearly $1,000$ experts representing over $500$ institutions across $50$ countries.
It has attracted considerable attention and is now regarded as one of the most challenging benchmarks for evaluating current state-of-the-art models.
MedReason~\cite{wu2025medreason} specifically evaluated a fine-grained medical subset of HLE, referred to as HLE (Med), which is distinct from the B/M subset introduced in~\citet{phan2025humanity}.
This medical subset provides new opportunities for detailed comparisons in terms of dataset statistics and evaluation difficulty.

Notably, \benchmark demonstrates a higher level of challenge compared to HLE (Med) as shown in~\citet{wu2025medreason}. \benchmark is not only more comprehensive but also presents significantly greater difficulty.
Despite the extensive human effort involved in the construction of HLE (Med), \benchmark is approximately $\mathbf{24}$x larger and imposes a more demanding evaluation.
These characteristics position \benchmark as the most comprehensive and challenging medical question-answering benchmark currently available.

\section{Additional Implementation Details}
\label{appx:implementation_details}

\subsection{AI Expert Models}
\label{appx:ai_models}
Specifically, the basic AI experts include \texttt{Qwen2.5-7B}~\cite{qwen2.5}
, \texttt{InternLM2.5-7B}~\cite{cai2024internlm2}, 
and \texttt{LLaMA-3.1-8B}~\cite{dubey2024llama} for text-only medical questions filtration, \texttt{Qwen2-VL-7B}~\cite{Qwen2VL}, \texttt{InternVL2-8B}~\cite{chen2024internvl}, 
and \texttt{LLaMA-3.2-11B-Vision}~\cite{dubey2024llama} for multimodal medical questions filtration.
The advanced AI experts include proprietary models \texttt{gpt-4o-2024-08-06} and \texttt{claude-3-5-sonnet-20241022}.

\subsection{Evaluation Settings}
\label{appx:evaluation_settings}

For both \texttt{o1} and \texttt{DeepSeek-R1}, we follow the corresponding recommended prompting guidelines to remove the system prompt.
For \texttt{DeepSeek-R1}, we also include the prompt: \textit{Put your final answer within \textbackslash$boxed\{\}$}.
Additionally, the API parameter \texttt{temperature} for these two models is unsupported.
For \texttt{QVQ-72B-Preview}, since it only supports single-round conversation, we remove the second round of zero-shot CoT and adjust the answer extraction script to handle choices within \textbackslash$boxed\{\}$.

\section{Error Analysis}

In this section, we analyze the reasons behind model errors by classifying different models' incorrect answers into several error types. We consider the following error types:

\begin{tcolorbox}[colback=lightblue!80,breakable,colframe=black]
\begin{itemize}[leftmargin=1em]
    \item \textbf{\textit{Reasoning Process Error:}} The model's prediction rationale indicates errors in key reasoning steps, which led to the incorrect answer.
    \item \textbf{\textit{Perceptual Error (for \benchmarkmm only):}} The incorrect answer stems from a misunderstanding or misinterpretation of the image or images provided in the question.
    \item \textbf{\textit{Question Understanding Error:}} The answer shows an incorrect understanding of the original question.
    \item \textbf{\textit{Lack of Medical Knowledge:}} The model's prediction shows a lack of medical knowledge necessary for arriving at the correct answer.
    \item \textbf{\textit{Formatting Error:}} The answer includes the correct content but is formatted improperly and causes the answer extraction process to fail.
\end{itemize}
\end{tcolorbox}

For each model analyzed below, we sample $200$ incorrectly answered questions from the \benchmarktext and \benchmarkmm subsets, respectively. We use \texttt{gpt-4o-2024-11-20} to label each error type based on the question, correct answer, correct explanation, incorrect answer, and incorrect prediction rationale produced by the model.

\begin{figure*}[ht]
    \centering
    \includegraphics[width=\linewidth]{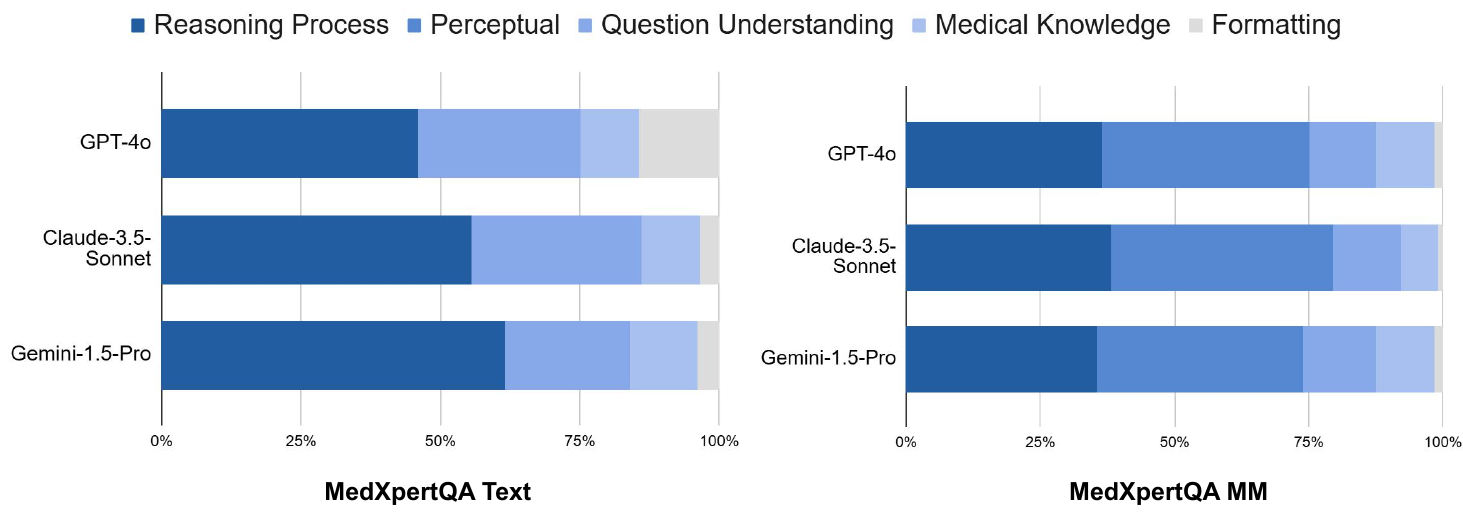}
    \caption{Error type distribution of three models on \benchmark.}
    \label{fig:error-analysis}
\end{figure*}

Figure~\ref{fig:error-analysis} compares error type distributions across the following models: \texttt{gpt-4o-2024-11-20}, \texttt{claude-3-5-sonnet-20241022}, \texttt{gemini-1.5-pro}.
Across all models, the high percentage of \textit{Reasoning Process Errors} on both \benchmarktext and \benchmarkmm reflects the difficulty of our benchmark in terms of medical reasoning. \textit{Perceptual Errors} are also prominent in \benchmarkmm, demonstrating not only room for improvement in medical image perception for leading models, but also that image interpretation is a core part of answering \benchmarkmm questions, as expected.

\textbf{In Appendix~\ref{appx:error-cases}, we provided examples of $4$ error types with detailed analysis.}

\section{Expert Review}
\label{appx:expert-review}

We summarize the error types found and corrected during the expert review of augmented questions and options. Table~\ref{tab:error-stats} provides a statistical overview.

\begin{table}[ht]
    \centering
    \begin{tabular}{ccc}
        \toprule
        \textbf{Subset} & \textbf{Error Type} & \textbf{\# Questions} \\
        \midrule
        \multirow{6}{*}{\benchmarktext}
            & Question Formatting Error & $15$ \\
            & Narrative Inconsistency & $5$ \\
            & Information Inconsistency & $73$ \\
            & Option Formatting Error & $21$ \\
            & Inapplicable Options & $399$ \\
            & Unreasonable Options & $15$ \\
        \midrule
        \multirow{4}{*}{\benchmarkmm}
            & Question Formatting Error & $100$ \\
            & Contained Question Source Information & $53$ \\
            & Narrative Inconsistency & $70$ \\
            & Information Inconsistency & $217$ \\
        \bottomrule
    \end{tabular}
    \caption{Statistics of errors identified during the human expert review process.}
    \label{tab:error-stats}
\end{table}

Appendix~\ref{appx:expert-review-guideline} lists reviewer guidelines for each phase of expert review in our benchmark construction and annotation.

\section{Full Results of Performance Comparison}\label{appx:full_performance_comparison}

Table~\ref{tab:full_performance_comparison} presents the results of different models on \benchmarktext and other text benchmarks.

\begin{table}[!ht]
    \centering
    \begin{tabular}{@{}lcccc@{}}
    \toprule
    Benchmark       & Qwen2.5-72B   & GPT-4o   & o1   & DeepSeek-R1 \\ \midrule
    MMLU (Medical)  & $89.62$       & $91.90^\ddag$   & $95.30^\ddag$  & $94.12$        \\
    MedQA-USMLE     & $72.70^\dag$ & $84.40^\ddag$   & $96.00^\ddag$ & $92.38$        \\
    MedMCQA-Dev     & $66.20^\dag$ & $76.40^\ddag$   & $83.90^\ddag$  & $79.12$        \\
    \benchmarktext & $18.90$       & $30.37$  & $44.67$ & $37.76$        \\ \bottomrule
    \end{tabular}
    \caption{Performance of different models on \benchmarktext and other benchmarks.
    $^\dag$ indicates the result is derived from~\citet{chen2024huatuogpt}.
    $^\ddag$ indicates the result is derived from~\citet{nori2024medprompt}.
    }
    \label{tab:full_performance_comparison}
\end{table}

\newpage

\section{Cases}
\label{appx:cases}

\tcbset{
    colback=white, 
    colframe=black, 
    width=\textwidth,
    arc=0mm          
}

\subsection{Question Cases}
\label{appx:question-cases}

\tcbset{
    width=474.18663pt,
    top=10pt,
    colback=white,
    colframe=black,
    colbacktitle=black,
    enhanced,
    center,
    attach boxed title to top left={yshift=-0.1in,xshift=0.15in},
    boxed title style={boxrule=0pt,colframe=white,},
}

\begin{center}
\textbf{\thesubsection.1 Example of MedQA}
\end{center}
\begin{tcolorbox}[title=Reasoning, colback=white, colframe=black, width=\textwidth, breakable]

\textbf{Question}: A 3-year-old boy presents to the emergency department with a ‘cough-like-a-seal bark’ and a high-pitched inspiratory noise that is audible without a stethoscope. His mother reports that his cough has worsened over the last few hours. The patient's blood pressure is 118/78 mm Hg, pulse is 90/min, respiratory rate is 35/min, and temperature is 38.3°C (101.1°F). On physical examination, the boy is sitting and leaning forward in apparent respiratory distress with suprasternal and intercostal retractions. Auscultation reveals inspiratory stridor without wheezing. He has a frequent barking cough and a hoarse voice when he speaks. What is a chest X-ray likely to show?

\textbf{Answer Choices:} (A) Increased interstitial markings (B) Lobar consolidation in the lingual (C) Thumbprint sign on the lateral image (D) Steeple sign

\end{tcolorbox}

\begin{tcolorbox}[title=Understanding, colback=white, colframe=black, width=\textwidth, breakable]

\textbf{Question}: A 3-month-old boy is brought the emergency department by his parents after an episode of cyanosis and muscle hypotonia that resolved after 2 minutes. Diagnostic evaluation fails to discover an exact etiology of the boy's symptoms and the episode is classified as a brief resolved unexplained event (BRUE). The risk profile for BRUE in infants remains largely unknown. The pediatrician who saw the boy in the emergency department is trying to identify risk factors for BRUE. She is aware of several confounders, including age, socioeconomic background, and family history of medical illness. She recruits 75 infants under 1 year of age with BRUE and 75 infants without BRUE of the same age, socioeconomic background, and family history of medical illness. She then compares the two groups with regard to history of feeding problems and history of recent upper respiratory infection. Which of the following methods was conducted to control confounding bias in the study?

\textbf{Answer Choices:} (A) Blinding (B) Restriction (C) Randomization (D) Matching

\end{tcolorbox}

\begin{center}
\textbf{\thesubsection.2 Example of \benchmarkmm}
\end{center}
\begin{tcolorbox}[title=Reasoning, colback=white, colframe=black, width=\textwidth, breakable]

\begin{wrapfigure}{R}{0.5\linewidth}
    \centering
    \includegraphics[width=\linewidth]{case_image_reasoning.pdf}
\end{wrapfigure}

\textbf{Question}: A 30-year-old male presents to primary care with complaints of gradually worsening fatigue and shortness of breath over several months. His medical history reveals no smoking, alcohol use, or illicit drug use, and his family history is noncontributory. He reports travel to South America one year prior but cannot recall any illness afterward. His vital signs show a temperature of 37.0$^\circ$C (98.6$^\circ$F), heart rate of 75/min, and blood pressure of 131/80 mmHg. Clinical examination reveals visible respiratory distress, a systolic murmur at the cardiac apex, and bilateral diffuse rales on lung auscultation. A cardiac biopsy is performed. Based on the biopsy findings shown, what is the most probable cause of the patient's condition?

\textbf{Answer Choices:} (A) Autoimmune granulomatous disease (B) Viral infection (C) Bacterial infection (D) Fungal infection (E) Parasitic infection

\end{tcolorbox}

\begin{tcolorbox}[title=Understanding, colback=white, colframe=black, width=\textwidth, breakable]

\begin{wrapfigure}{R}{0.5\linewidth}
    \centering
    \vspace{-0.4cm}
    \includegraphics[width=0.6\linewidth]{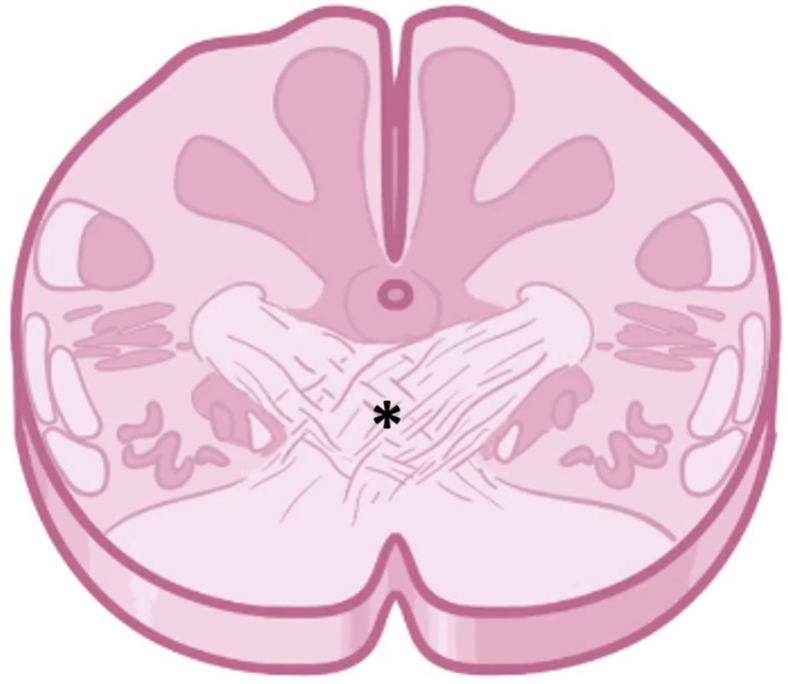}
\end{wrapfigure}

\textbf{Question}: A pathologist views cross-sectional slides of the brainstem and identifies the following structure (denoted by an asterisk).

Which of the following best characterizes the function of the fibers passing through this structure?

\textbf{Answer Choices:} (A) Transmission of pain signals to the brain (B) Transmission of unconscious proprioceptive sensation (C) Conduction of temperature sensation (D) Initiation of voluntary arm movements (E) Regulation of muscles of mastication

\end{tcolorbox}

\begin{center}
\textbf{\thesubsection.3 Example of \benchmarktext}
\end{center}
\begin{tcolorbox}[title=Reasoning, colback=white, colframe=black, width=\textwidth, breakable]

\textbf{Question}: A 68-year-old right-handed male is brought to the emergency department for evaluation of sudden-onset weakness. He woke up with clumsiness in his left hand, which quickly progressed to weakness on his left side, more pronounced in the arm than the leg. His medical history includes hypertension, type 2 diabetes mellitus, and hyperlipidemia, but he has not refilled his medications in a year. He has a 40-year history of smoking one pack of cigarettes daily. Vital signs reveal a temperature of 37.3$^\circ$C (99.1$^\circ$F), pulse of 98/min, respirations of 20/min, and blood pressure of 200/110 mm Hg. Physical examination shows slurred speech, left-sided facial weakness sparing the forehead, 2/5 strength in the left upper limb, and 4/5 strength in the left lower limb, with normal strength and sensation on the right side. MRI reveals a right-sided infarct in the paramedian pontine base. Which additional finding is most likely in this patient?

\textbf{Answer Choices:} (A) Ipsilateral tongue deviation (B) Contralateral loss of proprioception (C) Contralateral hemiparesis sparing the face (D) Eye deviation to the left (E) Vertical gaze palsy (F) Ipsilateral facial weakness involving the forehead (G) Eye deviation to the right (H) Loss of pupillary light reflex on the right (I) Bitemporal hemianopia (J) Contralateral hemianopia with macular sparing

\end{tcolorbox}

\begin{tcolorbox}[title=Understanding, colback=white, colframe=black, width=\textwidth, breakable]

\textbf{Question}: A 25-year-old previously healthy man presents to the clinic with cervical pain and stiffness that developed after a long-duration flight from California to New York the previous day. His symptoms have progressively worsened since onset. He reports no trauma to the area and denies other symptoms. Physical examination reveals a patient in mild discomfort with normal vital signs and no fever. The paraspinal region from C5-C7 demonstrates hypertonicity. Which motion represents the primary movement of the affected cervical segments?

\textbf{Answer Choices:} (A) Coupled rotation (B) Extension (C) Flexion (D) Sidebending (E) Lateral flexion (F) Axial compression (G) Rotation (H) Retraction (I) Translation (J) Protraction

\end{tcolorbox}

\newpage

\subsection{Error Cases}
\label{appx:error-cases}

\begin{tcolorbox}[title=Reasoning Process Error, colback=white, colframe=black, width=\textwidth, breakable]

\begin{wrapfigure}{R}{0.6\linewidth}
    \centering
    \includegraphics[width=1.0\linewidth]{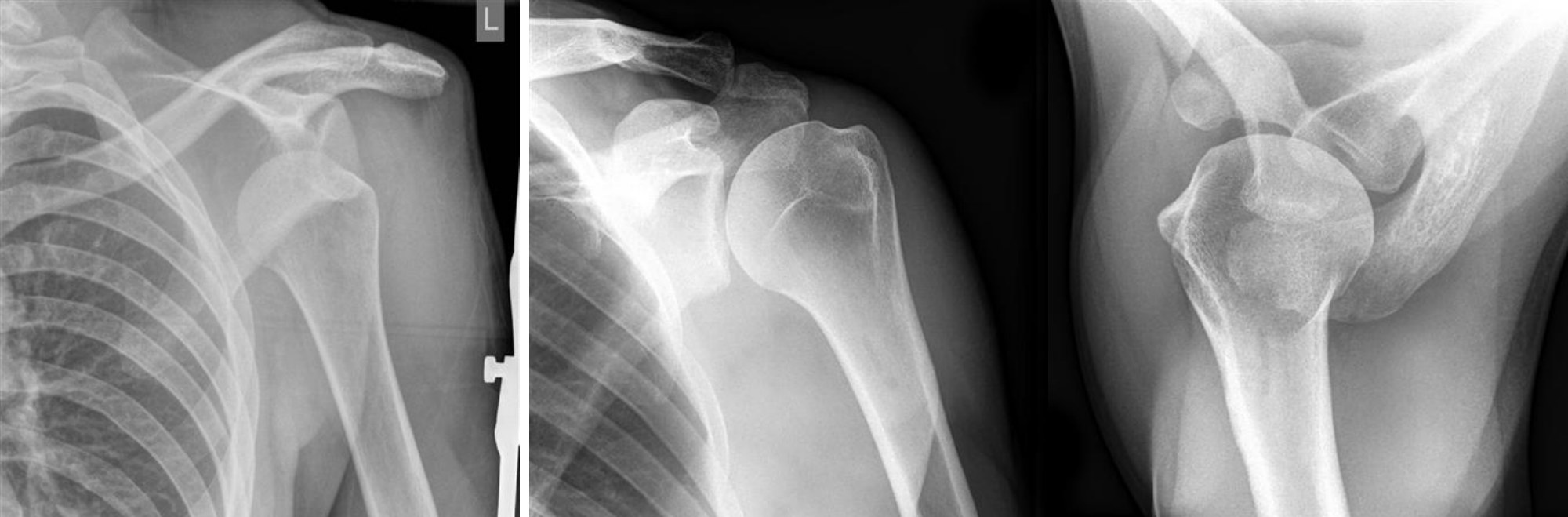}
\end{wrapfigure}

\textbf{Question}: A 64-year-old man is brought to the emergency room following a fall at his farm, presenting with a shoulder injury shown in Figure A. After successful reduction, post-reduction images are obtained (Figures B and C). During his follow-up appointment 10 days later, he demonstrates an inability to perform shoulder abduction. Which diagnostic test would be most appropriate to confirm the suspected diagnosis?

\textbf{Answer Choices:} (A) CT-angiogram of the affected extremity (B) EMG (C) MRI of the brachial plexus (D) MRI of the shoulder (E) Repeat shoulder x-rays

\textbf{Correct Answer:} \textcolor{green!70!black!70}{(D)}

\textbf{Model Answer:} \textcolor{red!70!black!70}{(B)}

\textbf{Prediction Rationale:} 

The presented scenario involves a 64-year-old man who sustained a shoulder injury after a fall and subsequently shows an inability to perform shoulder abduction following the incident. This could suggest nerve injury, specifically to the axillary nerve, which is responsible for innervating the deltoid muscle, crucial for shoulder abduction.

Here's the step-by-step breakdown:

1. Injury and Reduction: The man had a shoulder injury, likely a dislocation, based on initial x-rays and description. The reduction was successful, as indicated by follow-up x-rays.

2. Symptoms Post-Reduction: After 10 days, the patient demonstrates an inability to abduct the shoulder.

3. Common Cause of Inability to Abduct: Injury to the axillary nerve can cause an inability to abduct the shoulder, as it innervates the deltoid muscle.

4. Most Appropriate Diagnostic Test: Electromyography (EMG) is often used to assess the function of muscles and nerves. If there is a suspicion of nerve injury, such as the axillary nerve in this case, an EMG would be helpful in confirming nerve dysfunction or damage.

Considering the options: (A) CT-angiogram of the affected extremity would assess vascular issues, not nerve. (B) EMG would assess nerve injury. (C) MRI of the brachial plexus could be used for structural visualization of the nerve, but isn't typically the first choice for function assessment. (D) MRI of the shoulder focuses on soft tissue and structural injury, not specific nerve function. (E) Repeat shoulder x-rays assess bone placement, not nerve or muscle function.

Based on this analysis, the most appropriate diagnostic test to confirm the suspected axillary nerve injury would be (B) EMG.

\noindent\rule{\linewidth}{0.4pt}

\textbf{Analysis:} 

\textit{Upon approaching this question, the model correctly analyzed the provided X-ray images to determine that reduction for the patient had been successful and that X-rays showed no bone injuries. However, it proceeded to conclude that injury to the axillary nerve must have caused the patient's inability to abduct the shoulder, and followed this reasoning process to select the corresponding test, \textit{(B) EMG}. Although the model was correct in stating that axillary nerve injuries can impact the deltoid muscle, demonstrating correct medical knowledge, it does not sufficiently compare the likelihoods of several candidate diagnoses. Axillary nerve injury is relatively unlikely to fully account for the severe shoulder injury reported, and in comparison, a rotator cutoff tear is likely to occur in this demographic, and also better explains the patient's symptoms. This conclusion leads to the correct diagnostic test, \textit{(D) MRI of the shoulder}. This case reveals that the model followed the first plausible explanation it found and could not perform the nuanced reasoning necessary for weighing several possibilities. Multi-step reasoning is a notable challenge in this question - the answerer needs to first determine the most likely hypothetical diagnosis, then select the medical examination accordingly. The candidate diagnoses not being explicitly listed likely contributed to the model's oversimplified reasoning process.}

\end{tcolorbox}

\begin{tcolorbox}[title=Perceptual Error, colback=white, colframe=black, width=\textwidth, breakable]

\begin{wrapfigure}{R}{0.5\linewidth}
    \centering
    \includegraphics[width=1.0\linewidth]{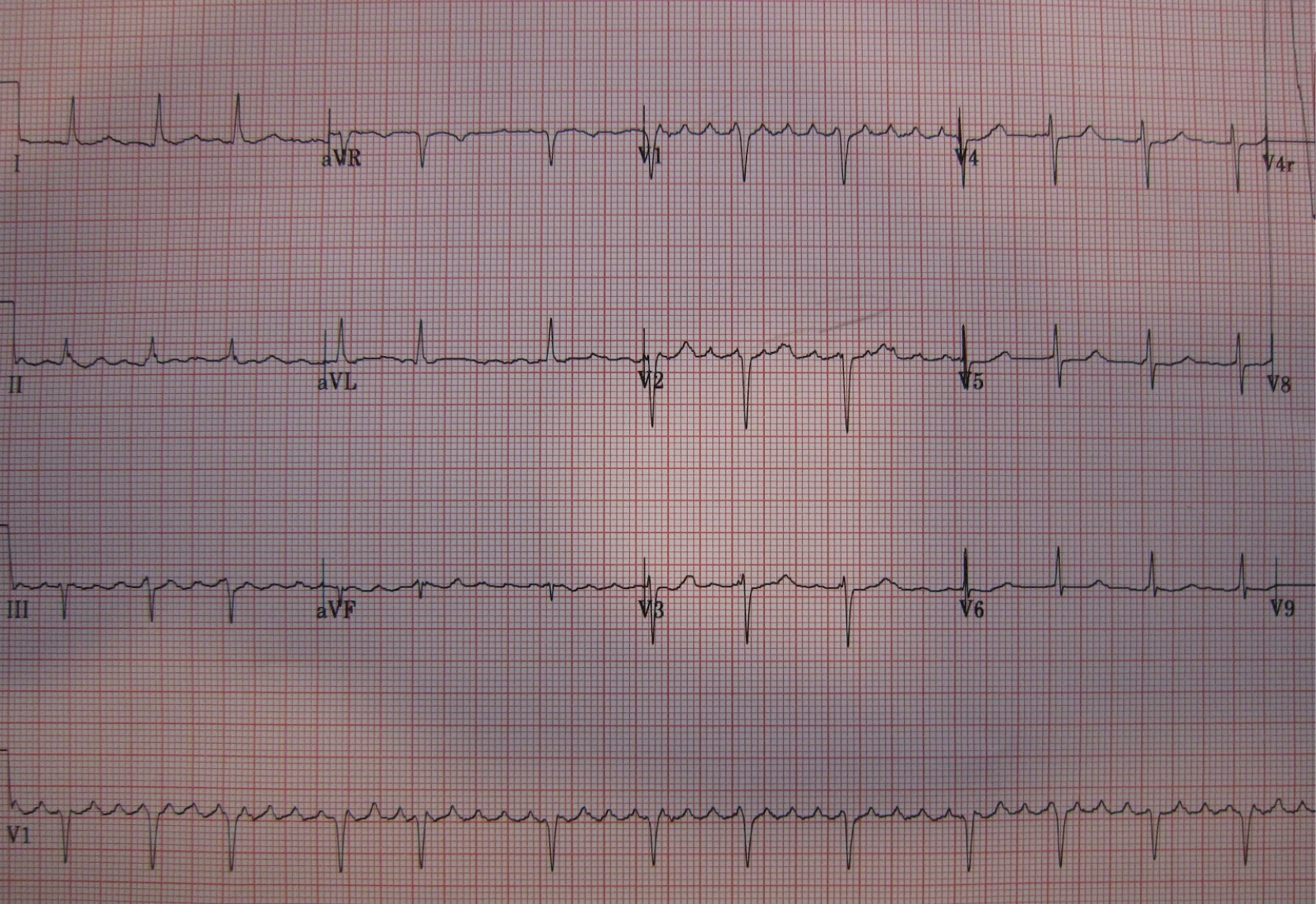}
\end{wrapfigure}

\textbf{Question}: A 57-year-old man is being monitored in the cardiothoracic intensive care unit following an uncomplicated coronary artery bypass graft surgery. During overnight monitoring, the patient's telemetry alerts staff to a tachyarrhythmia, prompting a rapid response call. His vital signs show: temperature 37.0°C (98.6°F), heart rate 140/min, respiratory rate 14/min, and blood pressure 134/72 mmHg. The patient reports only post-surgical chest pain at the incision site. Given the electrocardiogram findings shown below and the patient's clinical presentation, what is the most probable diagnosis?

\textbf{Answer Choices:} (A) Atrial flutter (B) Multifocal atrial tachycardia (C) Atrial fibrillation with rapid ventricular response (D) Paroxysmal supraventricular tachycardia (E) Sinus tachycardia

\textbf{Correct Answer:} \textcolor{green!70!black!70}{(A)}

\textbf{Model Answer:} \textcolor{red!70!black!70}{(C)}

\textbf{Prediction Rationale:} 

To determine the most probable diagnosis, we need to assess the clinical information alongside the ECG findings.

1. Heart Rate: The patient has a heart rate of 140 beats per minute, which indicates tachycardia.

2. ECG Analysis: The ECG shows an irregularly irregular rhythm. There are no distinct P waves visible before each QRS complex; instead, there is a disorganized electrical activity, which is typical of fibrillatory waves.

3. Differential Diagnosis: (A) Atrial flutter: Typically gives a ``saw-tooth'' pattern and usually has a more regular rhythm. (B) Multifocal atrial tachycardia (MAT): Characterized by an irregular rhythm with at least three different P wave morphologies which isn't apparent here. (C) Atrial fibrillation with rapid ventricular response: Characterized by an irregularly irregular rhythm without distinct P waves, which fits the ECG findings. (D) Paroxysmal supraventricular tachycardia (PSVT): Usually presents with a regular rhythm and wouldn't have the irregular characteristics seen in this case. (E) Sinus tachycardia: Would have a regular rhythm and identifiable P waves.

4. Post-Surgical Context: Atrial fibrillation (AF) is relatively common after cardiac surgery, such as coronary artery bypass grafting, due to inflammation, electrolyte imbalances, and other stressors on the heart.

Considering the irregularly irregular rhythm and absence of distinct P waves, the most probable diagnosis is: (C) Atrial fibrillation with rapid ventricular response.

\noindent\rule{\linewidth}{0.4pt}

\textbf{Analysis:} 

\textit{The provided ECG demonstrates characteristic flutter waves with a 3:1 and 4:1 AV nodal block, indicative of \textit{(A) atrial flutter}. The model correctly states that atrial flutter typically gives a ``saw-tooth'' pattern, but fails to determine that such patterns are present in the image. Instead, it incorrectly interprets the ECG as having an irregularly irregular rhythm indicative of \textit{(C) atrial fibrillation}. The main distinction between these two diagnoses is the regularity of atrial activations seen in the ECG, a detailed piece of visual information that is challenging for models to accurately interpret.}

\end{tcolorbox}

\newpage

\begin{tcolorbox}[title=Question Understanding Error, colback=white, colframe=black, width=\textwidth, breakable]

\textbf{Question:} A 60-year-old male with multiple comorbidities including coronary artery disease (status post stent placement 3 months ago), hypertension, diabetes, and mild to moderate asthma presents with a two-day history of cough and wheeze. His medication regimen includes aspirin, atorvastatin, atenolol, metformin, and hydrochlorothiazide. He recently added salbutamol inhaler and prednisolone for his respiratory symptoms. The patient reports experiencing occasional low blood glucose readings over the past few weeks. Physical examination reveals stable vital signs with no respiratory distress. Which medication in his current regimen is the most likely cause of his hypoglycemic episodes?

\textbf{Answer Choices:} (A) Aspirin (B) Ibuprofen (C) Enalapril (D) Naproxen (E) Atorvastatin (F) Prednisolone (G) Atenolol (H) Hydrochlorothiazide (I) Phenformin (J) Metformin

\textbf{Correct Answer:} \textcolor{green!70!black!70}{(A)}

\textbf{Model Answer:} \textcolor{red!70!black!70}{(G)}

\textbf{Prediction Rationale:}

When analyzing this case, we need to identify which medication in the patient’s regimen is the most likely cause of his hypoglycemic episodes. Here is a step-by-step approach:

Step 1: Review the patient’s medications: 1. Aspirin – Antiplatelet therapy. 2. Atorvastatin – Lipid-lowering agent (statin). 3. Atenolol – A beta-blocker used for hypertension. 4. Metformin – A first-line diabetes medication. 5. Hydrochlorothiazide – A diuretic used for blood pressure control. 6. Salbutamol (recently added) – A short-acting beta-agonist for asthma. 7. Prednisolone (recently added) – A corticosteroid for respiratory inflammatory issues.

Step 2: Focus on the patient’s problem – hypoglycemia: The patient reports low blood glucose readings. Hypoglycemia can be caused by certain medications, especially those that:  1. Lower blood glucose directly (e.g., diabetes medications). 2. Affect glucose metabolism indirectly.

Step 3: Evaluate each medication for its relationship to hypoglycemia

- Aspirin: At high doses, it can rarely cause hypoglycemia. However, at low doses (commonly used for antiplatelet therapy), this is unlikely. Not the most likely cause.

- Ibuprofen and Naproxen (if taken, not explicitly listed in the regimen): NSAIDs may not be directly relevant to hypoglycemia unless used at high doses affecting insulin function. Likely not relevant here.

- Enalapril: ACE inhibitors can rarely cause hypoglycemia by increasing insulin sensitivity. However, this medication is not listed in his current regimen. Irrelevant.

- Metformin: Unlikely to cause hypoglycemia as it does not stimulate insulin secretion. Its primary effect is reducing hepatic gluconeogenesis. Not the cause.

- Atorvastatin: Statins are not associated with significant risks of hypoglycemia. Not the cause.

- Prednisolone: Corticosteroids typically increase blood glucose (hyperglycemia), not decrease it. Not the cause.

- Atenolol: Beta-blockers, particularly non-cardioselective ones, can mask the symptoms of hypoglycemia (e.g., tachycardia) and impair the response to hypoglycemia (e.g., by reducing glycogenolysis). Cardioselective beta-blockers, such as atenolol, are less likely to do this but can still blunt the counterregulatory response during hypoglycemia. Possible contributor.

- Hydrochlorothiazide: Thiazides are more typically associated with hyperglycemia. Unlikely to cause hypoglycemia.

Step 4: Consider the most likely cause: The most likely cause of the hypoglycemia is Atenolol because: 1. Beta-blockers impair the body’s natural response to a drop in glucose by reducing glycogenolysis and gluconeogenesis. 2. They also mask typical warning signs of hypoglycemia (e.g., palpitations or tremors), which may lead to unrecognized or prolonged hypoglycemic episodes.

\noindent\rule{\linewidth}{0.4pt}

\textbf{Analysis:} 

\textit{The above response fails to fully grasp the question's implications in several ways. For the incorrect answer \textit{(G) Atenolol}, the model generally states that some beta-blockers reduce glycogenolysis and gluconeogenesis, but does not consider the specificity of atenolol - as a selective beta-1 blocker, it has less effect on glucose metabolism. While the claim that it masks warning signs of hypoglycemia is more accurate, this does not align with the question's emphasis on finding a \textit{cause} of hypoglycemia. For the correct answer \textit{(A) Aspirin}, the model's judgment is only accurate regarding non-diabetic individuals, while ample evidence supports aspirin's role in decreasing glucose levels in diabetic patients. The remaining parts of the model's analysis similarly disregard the context of the patient's diabetes. The response is therefore inconsistent with the patient's general background presented in the question.
}

\end{tcolorbox}

\newpage

\begin{tcolorbox}[title=Lack of Medical Knowledge, colback=white, colframe=black, width=\textwidth, breakable]

\textbf{Question}: A 69-year-old male visits his primary care physician for a routine checkup. He maintains an active lifestyle since retirement, engaging in daily exercise and home projects, and follows a balanced diet with regular evening time with his spouse. However, he reports feeling less refreshed upon waking compared to his younger years, when he slept 9 hours and felt excellent, versus his current 7 hours of sleep. His wife reports that his sleep appears peaceful. His medical conditions include hypertension and diabetes, managed with lisinopril and metformin. Physical examination reveals: temperature 98.6°F (37°C), blood pressure 131/85 mmHg, pulse 71/min, respirations 12/min, normal neurological findings, good muscle tone, healthy weight, and pleasant affect. He denies current fatigue or tiredness. Which physiological alteration would most likely be observed in this patient?

\textbf{Answer Choices:} (A) Reduced acetylcholine (B) Increased serotonin (C) Decreased orexin (D) Altered circadian rhythm proteins (E) Elevated histamine levels (F) Decreased melatonin (G) Elevated cortisol levels (H) Increased dopamine (I) Decreased growth hormone (J) Increased adenosine

\textbf{Correct Answer:} \textcolor{green!70!black!70}{(F)}

\textbf{Model Answer:} \textcolor{red!70!black!70}{(I)}

\textbf{Prediction Rationale:} 

This scenario describes an older adult experiencing reduced restorative sleep, which is a common part of the aging process. To determine the most likely physiological alteration, it is important to consider how aging impacts sleep.

Key details from the case:

1. Age-related sleep changes: The patient is 69 years old. Aging is associated with alterations in sleep architecture, which may lead to reductions in total sleep duration, slow-wave (deep) sleep, and sleep efficiency. Older adults often feel less refreshed upon waking, even when sleep appears otherwise normal.

2. Patient's sleep specifics: He now sleeps 7 hours and wakes feeling less refreshed compared to his younger years when sleeping 9 hours. His sleep appears peaceful, and he experiences no disruptive symptoms such as snoring or awakenings.

3. Ruling out other causes: Good management of hypertension and diabetes, normal vitals, a well-rounded active lifestyle, and no reported fatigue or significant tiredness suggest no acute medical condition contributing to his symptoms.

4. Physiological changes associated with aging: Aging often leads to a decrease in growth hormone, which is secreted primarily during slow-wave sleep and stimulates restorative processes during sleep. This can result in less restorative sleep and a less refreshed feeling upon waking.

5. Other options to consider: Decreased melatonin (option F): Melatonin levels decline with age, but melatonin primarily regulates sleep initiation and circadian rhythm rather than sleep quality or restoration. Altered circadian rhythm proteins (option D): Circadian changes occur with aging, but his symptoms are more closely linked to sleep restoration rather than timing. Other neurotransmitters like acetylcholine, orexin, serotonin, histamine, cortisol, and adenosine are not directly relevant to the described problem.

Correct Answer: (I) Decreased growth hormone. This is the most likely physiological change contributing to the patient’s symptoms, as reduced growth hormone secretion in older adults is linked to decreased restorative properties of sleep.

\noindent\rule{\linewidth}{0.4pt}

\textbf{Analysis:} 

\textit{The question generally describes a patient experiencing reduced sleep quality, commonly associated with aging. The model's response reflects its insufficient medical knowledge on the different effects of \textit{(D) decreased growth hormone} and \textit{(F) decreased melatonin} on sleep. While both are indeed correlated with aging, \textit{growth hormones} primarily promotes deep sleep, or slow-wave sleep (SWS), while \textit{decreased melatonin} more directly accounts for the subjective lack of restfulness upon waking referred to in the question.}

\end{tcolorbox}

\newpage

\subsection{Evaluation Case}
\label{appx:evaluation-case}

\begin{tcolorbox}[colback=white, colframe=black, width=\textwidth, breakable]

\textbf{System:}
You are a helpful medical assistant.

\textbf{User:}
Q: A 57-year-old male presents to the emergency department with a one-week history of weakness and subjective fever. His medical history includes IV drug use, hepatitis C, atrial fibrillation, cirrhosis, alcohol dependence, obesity, and depression. On examination, his temperature is 102°F (38.9°C), blood pressure is 157/98 mmHg, pulse is 110/min, respirations are 17/min, and oxygen saturation is 98\% on room air. The patient appears fatigued with diffuse jaundice. Cardiopulmonary examination reveals bibasilar crackles, and abdominal examination shows distension, dullness to percussion, a fluid wave, and generalized tenderness. The patient is admitted to the hospital and started on piperacillin-tazobactam. By day 4, he is afebrile, with a pulse of 92/min, reduced abdominal tenderness, and persistent jaundice. Nurses report bleeding from two peripheral IV sites requiring pressure to control, along with new bruises on the patient's arms and legs. What is the most likely explanation for this patient's condition?

Answer Choices: (A) Fibrinolysis activation due to liver failure (B) Vitamin K deficiency due to liver dysfunction (C) Diffuse activation of the coagulation cascade (D) Disseminated intravascular coagulation (DIC) (E) Platelet dysfunction secondary to sepsis (F) Decreased production of coagulation factors (G) Immune-mediated platelet destruction (H) Hypersplenism causing thrombocytopenia (I) Decreased metabolism of an anticoagulant (J) Bacterial destruction

A: Let's think step by step.

\textbf{Assistant:}
Analysis:

This patient has advanced liver disease (cirrhosis) and is hospitalized on broad‐spectrum antibiotics for four days. Although his infection is improving (afebrile, lower heart rate, less abdominal tenderness), he has developed new signs of a bleeding diathesis (bleeding from IV sites, new bruises). In a patient with chronic liver disease, an “acute on chronic” coagulopathy like this most often points to additional impairment in clotting‐factor synthesis, above and beyond what is already chronically reduced by cirrhosis.

Two common acute drivers of coagulopathy in cirrhotic patients are:  

\begin{itemize}
    \item Vitamin K deficiency (especially when on broad‐spectrum antibiotics that eradicate gut flora responsible for synthesizing vitamin K).

    \item Disseminated intravascular coagulation (DIC), usually in the setting of sepsis or shock, but here the patient’s sepsis is improving, making DIC less likely.
\end{itemize} 

Thus, the best explanation in a cirrhotic patient, newly on broad‐spectrum antibiotics, who suddenly develops worse bleeding tendencies, is vitamin K deficiency superimposed on his baseline liver dysfunction.

Answer: (B) Vitamin K deficiency due to liver dysfunction.

\textbf{User:}
Therefore, among A through J, the answer is

\textbf{Assistant:}
The correct answer is (B) Vitamin K deficiency due to liver dysfunction.
    
\end{tcolorbox}

\section{Expert Review Guideline}
\label{appx:expert-review-guideline}

\begin{table}[H]
\centering
\begin{tabular}{p{\textwidth}}
\toprule
\underline{\textbf{\textsc{Guidelines for Expert Review of Augmented Questions.}}} \\
\begin{minipage}{\textwidth}
\vspace{2mm}
\begin{minted}[fontsize=\small,breaklines=true]{markdown}

**Intended Audience:** Human medical expert reviewers
**Objective:** Conduct a comprehensive review and proofreading of medical questions rewritten by large language models to ensure that the questions are accurate, clear, scientific, practical, and free from hallucinations or factual errors.

## 1. Review of Original Question

**Objective:** Verify whether the original question contains any logical or factual errors, is incomplete, or unanswerable. If such issues exist, no further steps should be taken, and the specific situation should be recorded.

**Requirements:** Clearly indicate the problems in the original question. Mark the question as invalid.

## 2. Review of Rewritten Question

### 2.1 Content Accuracy Check

**Objective:** Ensure the rewritten question maintains medical knowledge accuracy and is free from factual errors.

**Key Points:** Check whether any key information from the original question is missing. Verify that no inaccurate or deviated content has been introduced in the rewrite. Ensure that no content contradicts medical facts or deviates from the original context.

### 2.2 Language Clarity and Expression Quality Check

**Objective:** The language of the rewritten question should be precise, clear, and professional.

**Key Points:** Ensure the language is professional and avoids overly casual or colloquial expressions. Confirm that medical terms are used accurately and scientifically, clearly conveying the meaning of the original question. Verify that the expression is concise and clear, avoiding ambiguity or unclear phrasing.

### 2.3 Logical Consistency Check

**Objective:** Ensure the question follows a logical flow, has a reasonable structure, and is answerable.

**Key Points:** Check whether the question fits into real-world clinical or research context. Look for logical errors such as contradictions, lack of causality, reasoning mistakes, or illogical phrasing. Ensure the question structure is clear, complete, coherent, and has a clear focus.

### 2.4 Other Issues

**Tablular Data:** If the original question includes tabular data, verify that the rewritten question retains the original data's content and format. The rewritten table should not omit any data or alter the format.

**Medical Terminology Abbreviations:** If the rewritten question replaces the full medical term with an abbreviation, verify whether the abbreviation is standard and in line with medical conventions. If professional medical practitioners can understand the abbreviation in the context of the question, it is considered acceptable.
\end{minted}
\end{minipage} \\
\bottomrule
\end{tabular}
\caption{Guidelines for expert review of augmented questions (Part 1).}
\label{tab:question-review-guidelines-1}
\end{table}
\newpage

\begin{table}[htbp]
\centering
\begin{tabular}{p{\textwidth}}
\toprule
\underline{\textbf{\textsc{Guidelines for Expert Review of Augmented Questions.}}} \\
\begin{minipage}{\textwidth}
\vspace{2mm}
\begin{minted}[fontsize=\small,breaklines=true]{markdown}

## 3. Summary of Requirements

**Strict Adherence to Standards:** All medical experts must strictly follow these guidelines for evaluation and proofreading to ensure the accuracy, scientific integrity, and practicality of the questions.

**Quality of Evaluation Results:** The content must be accurate, with reliable facts. The language expression should be clear and meet professional medical standards.
The logic should be sound, and the questions should be answerable.

**Maintain Objectivity:** Maintain objectivity and fairness throughout the evaluation and proofreading process, avoiding subjective judgment or misleading rewrites.

\end{minted}
\end{minipage} \\
\bottomrule
\end{tabular}
\caption{Guidelines for expert review of augmented questions (Part 2).}
\label{tab:question-review-guidelines-2}
\end{table}

\newpage

\begin{table}[H]
\centering
\begin{tabular}{p{\textwidth}}
\toprule
\underline{\textbf{\textsc{Guidelines for Expert Review of Augmented Options.}}} \\
\begin{minipage}{\textwidth}
\vspace{2mm}
\begin{minted}[fontsize=\small,breaklines=true]{markdown}

**Intended Audience:** Human medical expert reviewers

**Objective:** Conduct a comprehensive review and proofreading of the expanded options generated by large language models for medical multiple-choice questions, ensuring the options are accurate, clear, reasonable, and challenging.

## 1. Option Inspection Standards

### 1.1 Content Rationality Check

**Objective:** Ensure the expanded options are reasonable and acceptable within the context of the original question.

**Key Points:** Avoid adding options that do not align with the question's context. For example, if there are 4 images in the question, generating an option for “Image 5” would be unacceptable. Avoid adding options that do not correspond to the topic of the question. For example, if the question asks to choose from several time periods, do not generate options with unrelated content.

### 1.2 Language Expression and Professionalism Check

**Objective:** Ensure the language of the options is accurate, clear, and professional.

**Key Points:** The language style of the expanded options should match that of the original options. The language should be professional, avoiding overly casual or colloquial expressions. Medical terminology should be used accurately and in line with current medical standards. The expression should be clear, concise, and avoid ambiguity or overly complex phrasing.

### 1.3 Hallucination Check

**Objective:** Ensure that the original options do not contain obvious hallucinations or errors.

**Key Points:** The original options should not be changed.
The generated options should not contain hallucinated information, i.e., content that is disconnected from the original question context or lacks scientific basis. There should be no repetition of content between options, and no redundant information in the options.

## 4. Summary and Requirements

**Strict Adherence to Standards:** All medical experts must strictly follow these guidelines for evaluation and proofreading to ensure the accuracy, scientific integrity, and practicality of the options.

**Quality of Evaluation Results:** Content must be accurate and reliable. Language expression should be clear and meet professional medical standards. Option design should be logically sound and have a high degree of differentiation.

**Maintain Objectivity:** Maintain objectivity and fairness throughout the evaluation and proofreading process, avoiding subjective judgment or misleading expansions.

\end{minted}
\end{minipage} \\
\bottomrule
\end{tabular}
\caption{Guidelines for expert review of augmented options.}
\label{tab:option-review-guidelines}
\end{table}

\newpage

\section{Prompts}
\label{appx:prompts}

The following lists all prompts used in this work.

\subsection{Attribute Annotation Prompts}
\label{appx:attribute-annotation-prompts}

\begin{table}[H]
\centering
\begin{tabular}{p{\textwidth}}
\toprule
\underline{\textbf{\textsc{Prompt for Task Annotation.}}} \\
\begin{minipage}{\textwidth}
\vspace{2mm}
\begin{minted}[fontsize=\small,breaklines=true]{markdown}

You are an experienced medical doctor and independent practitioner. You will be asked to label a medical question according to its task.

You will be given a list of tasks and their corresponding explanations and descriptions. You will then be given a medical examination question.
Please determine which task the question best pertains to.

If the question is related to multiple tasks, only select the most relevant one.
Directly output the task name you have selected. Do not provide any additional explanations. Ensure that it is a valid task within the list of available tasks provided.

**Tasks:**

Diagnosis: Tasks that occur in the diagnosis stage of a medical examination. These include identifying diseases based on symptoms, deducing the causes of symptoms or diseases, predicting disease prognosis or additional symptoms, and proposing intermediate steps (such as further examinations) to facilitate diagnosis.

Treatment: Tasks that occur in the treatment stage of a medical examination. These include selecting treatments for diseases, proposing preventive measures for diseases, and suggesting lifestyle changes for the patient.

Basic Medicine: Tasks that focus on basic scientific principles that may be applied in medical practice. These include understanding anatomy, diseases and medical procedures, and general STEM principles.

**Question:**

{question_text}

**Output:**

\end{minted}
\end{minipage} \\
\bottomrule
\end{tabular}

\caption{Prompt for Task Annotation.}
\label{tab:task-annotation-prompt}
\end{table}
\newpage

\begin{table}[H]
\centering
\begin{tabular}{p{\textwidth}}
\toprule
\underline{\textbf{\textsc{Prompt for Subtask Annotation.}}} \\
\begin{minipage}{\textwidth}
\vspace{2mm}
\begin{minted}[fontsize=\small,breaklines=true]{markdown}

You are an experienced medical doctor and independent practitioner. You will be asked to label a medical question's subtask according to its task.

You will be given a medical examination question and its task type. You will then be given a list of subtypes within the predefined type. Please determine which subtask the question best pertains to.
If the question is related to multiple subtasks, only select the most relevant one. Directly output the subtask name you have selected. Do not provide any additional explanations. Ensure that it is a valid task within the list of available tasks provided.

**Task:** Diagnosis

**Subtasks:**

Primary, Differential, Etiologic, Prognostic, Retrospective, Syndromic, Predictive, Intermediate

**Question:**

{question_text}

**Output:**

\end{minted}
\end{minipage} \\
\bottomrule
\end{tabular}

\caption{Prompt for Subtask Annotation. We use the prompt for diagnosis tasks as an example. Each question's prompt will contain the specific subtasks within the question's labeled task.}
\label{tab:subtask-annotation-prompt}
\end{table}

\begin{table}[!htbp]
\centering
\begin{tabular}{p{\textwidth}}
\toprule
\underline{\textbf{\textsc{Prompt for System Annotation.}}} \\
\begin{minipage}{\textwidth}
\vspace{2mm}
\begin{minted}[fontsize=\small,breaklines=true]{markdown}

You are an experienced medical doctor and independent practitioner. Your task will be to label a medical question according to the human body system it corresponds to.

You will be given a list of human body systems, followed by a medical question. Please determine which system the question best pertains to. If the question is related to multiple systems, only select the most relevant one.
Directly output the name of the final system you selected from th list of available systems.

**Systems:**
Integumentary, Skeletal, Muscular, Nervous, Endocrine, Cardiovascular, Lymphatic, Respiratory, Digestive, Urinary, Reproductive

**Question:**

{question_text}

**Output:**

\end{minted}
\end{minipage} \\
\bottomrule
\end{tabular}

\caption{Prompt for System Annotation.}
\label{tab:system-annotation-prompt}
\end{table}

\begin{table}[H]
\centering
\begin{tabular}{p{\textwidth}}
\toprule
\underline{\textbf{\textsc{Prompt for Reasoning and Understanding Annotation.}}} \\
\begin{minipage}{\textwidth}
\vspace{2mm}
\begin{minted}[fontsize=\small,breaklines=true]{markdown}

You are an experienced medical doctor and independent practitioner. Your task will be to determine whether a medical question primarily challenges the answerer's medical knowledge understanding or medical reasoning skills.

You will be given a challenging medical exam question designed for professional medical practitioners, followed by its correct answer and explanation to help you understand the question.
Please determine whether the question primarily challenges the answerer's medical knowledge understanding or medical reasoning ability.
Consider the amount of information provided in the question, as well as the number and complexity of reasoning steps required to answer it.
You will be given the question you will label with its answer and explanation. Directly output either "Understanding" Or "Reasoning" as your answer without any additional information or explanations.

**Question:**

{question_text}

**Explanation:**

{explanation_text}

**Output:**

\end{minted}
\end{minipage} \\
\bottomrule
\end{tabular}

\caption{Prompt for annotating questions based on their focus on either reasoning or understanding skills, used for constructing a reasoning subset of \benchmark.}
\label{tab:reasoning-annotation-prompt}
\end{table}

\newpage

\subsection{Data Augmentation Prompts}
\label{appx:data-augmentation-prompts}

\begin{table}[H]
\centering
\begin{tabular}{p{\textwidth}}
\toprule
\underline{\textbf{\textsc{Prompt for Question Rephrase.}}} \\
\begin{minipage}{\textwidth}
\vspace{2mm}
\begin{minted}[fontsize=\small,breaklines=true]{markdown}
### Task

Rephrase a given medical multiple-choice question according to the following guidelines:

### Suggested Approaches

Consider the following strategies to rephrase the question:

1. **Language Rewriting**: Use different expressions and sentence structures that are equivalent in meaning to the original.

2. **Information Restructuring**: Alter the order that information is presented in the original question without disrupting the logical coherence of the question.

### General Requirements

1. **Maintain the core content:** Ensure that the rephrased question retains the core content, reasoning logic, and correct answer of the original question.

2. **Ensure a professional language style:** Maintain a professional, formal, and clear language style similar to the original question. Rigorously ensure clarity and avoid ambiguity. Feel free to copy parts of the original question if alternative appropriate phrasing is not possible.

3. **Maintain factual consistency:** Ensure that the rewritten question retains every piece of information in the original. Do not change, add, or delete any factual information.

4. **Imitate original formatting:** Keep any special formatting in the original question unchanged, especially regarding structured data presentation. Pay special attention to keep any tabular data in completely the same format as the original.

5. **Final output format:** Ensure that the options section of the question remains unchanged and the format remains as "Answer Choices: (A) [Option A] (B) [Option B] ...". Only output the rephrased question. Do not include any additional information or explanations.

{demonstrations}

### Input

**Original Question:**
{question}

**Correct Answer:**
{label}

**Explanation:**
The following is an explanation to help with understanding the question:
{explanation}

### Output

**Rephrased Question:**
\end{minted}
\end{minipage} \\
\bottomrule
\end{tabular}
\caption{Prompt for Question Rephrase.}
\label{tab:question-rephrase-prompt}
\end{table}
\newpage

\begin{table}[H]
\centering
\begin{tabular}{p{\textwidth}}
\toprule
\underline{\textbf{\textsc{Prompt for Option Generation.}}} \\
\begin{minipage}{\textwidth}
\vspace{2mm}
\begin{minted}[fontsize=\small,breaklines=true]{markdown}
### Task

Given a medical multiple-choice question designed to evaluate the capabilities of medical experts and its correct answer, please design additional incorrect options to challenge experts when they answer the question.

### Suggested Approaches

1. **Consider Errorneous Perspectives:** Include distractors that interpret key information in the question incorrectly.

2. **Leverage Common Misconceptions:** Consider designing distractors leveraging common errors or medical concepts that are frequently confused.

3. **Logical Misdirection:** Introduce distractors grounded in logical reasoning that is seemingly plausible but incorrect.

### General Requirements

1. **Maintain Consistency:** Ensure that the generated new options match the original options in terms of length, structure, word count, and grammatical form. They should be clear, concise, and professionally worded.

2. **Avoid Oversimplified Distractors:** Do not include options that can be easily dismissed based on intuition or surface-level analysis.

3. **Ensure High Plausibility:** Maintain the plausibility of each generated option. Avoid options that are overtly illogical or unsupported.

4. **Final Format:** Present the original question and options, followed by the **{generate_num}** additional options. Ensure that the generated options follow the same format as the original: "Answer Choices: (A) [Option A] (B) [Option B] ...". Do not output anything after the options.

{demonstrations}

### Input

**Original Question:**
{question}

**Correct Answer:**
{label}

### Output

**{generate_num}** options appended after the original question:
\end{minted}
\end{minipage} \\
\bottomrule
\end{tabular}

\caption{Prompt for Option Generation.}
\label{tab:option-generation-prompt}
\end{table}

\clearpage

\end{document}

%% file: inference_time_scaling.tex
\newcommand{\reasoning}{{\fontfamily{ppl}\selectfont Reasoning\;}}
\newcommand{\knowledge}{{\fontfamily{ppl}\selectfont Understanding\;}}
\begin{tikzpicture}  

    \node[draw, fill=red!40!black!15, minimum width=0.9em, minimum height=0.6em, label=right:\scriptsize{\texttt{DeepSeek-V3}}] 
    (d1) at (-23.4em, 17em) {};
    \node[draw, fill=red!90!black!70, minimum width=0.8em, minimum height=0.6em, anchor=north, 
    label=right:\scriptsize{\texttt{DeepSeek-R1}}] 
    (d2) at ([yshift=-0.5em]d1.south) {};
    
    \node[draw, fill=cyan!20!blue!40, minimum width=0.8em, minimum height=0.6em, 
    label=right:\scriptsize{\texttt{Qwen2.5-32B-Instruct}}] 
    (d1) at (-7.3em, 17em) {};
    \node[draw, fill=cyan!30!blue!70, minimum width=0.8em, minimum height=0.6em, anchor=north, 
    label=right:\scriptsize{\texttt{QwQ-32B-Preview}}] 
    (d2) at ([yshift=-0.5em]d1.south) {};
    
    \node[draw, fill=orange!40, minimum width=0.8em, minimum height=0.6em, label=right:\scriptsize{\texttt{Qwen2-VL-72B}}] 
    (d1) at (8.7em, 17em) {};
    \node[draw, fill=orange!85, minimum width=0.8em, minimum height=0.6em, anchor=north, 
    label=right:\scriptsize{\texttt{QVQ-72B-Preview}}] 
    (d2) at ([yshift=-0.5em]d1.south) {};

    \scriptsize{
        \begin{axis}[
            at={(-34em, 11em)},
            ybar,
            height=.24\textwidth,
            width=.29\textwidth,
            grid style=dashed,
            bar width=1.1em,
            tick align=inside,
            ylabel style={yshift=-1em},
            ylabel={\textbf{Accuracy}},
            symbolic x coords={{\knowledge,\reasoning}},
            enlarge x limits=0.55,
            xtick=data, 
            ymin=20, ymax=40,
            yticklabel style={/pgf/number format/precision=1, /pgf/number format/fixed zerofill}
        ]
            \addplot[fill=red!40!black!15, draw] coordinates { ({\reasoning}, 23.59) ({\knowledge}, 25.13)};
            \addplot[fill=red!90!black!70, draw] coordinates {({\reasoning}, 37.88) ({\knowledge}, 37.35)};
        \end{axis}
    }

    \scriptsize{
        \begin{axis}[
            at={(-11em, 11em)},
            ybar,
            height=.24\textwidth,
            width=.29\textwidth,
            grid style=dashed,
            bar width=1.1em,
            tick align=inside,
            ylabel style={yshift=-1em},
            symbolic x coords={{\knowledge,\reasoning}},
            enlarge x limits=0.55,
            xtick=data,
            ymin=12, ymax=20,
            yticklabel style={/pgf/number format/precision=1, /pgf/number format/fixed zerofill}
        ]
            \addplot[fill=cyan!20!blue!40, draw] coordinates { ({\reasoning}, 12.74) ({\knowledge}, 17.66)};
            \addplot[fill=cyan!30!blue!70, draw] coordinates {({\reasoning}, 18.86) ({\knowledge}, 15.11)};
        \end{axis}
    }

    \scriptsize{
        \begin{axis}[
            at={(12em, 11em)},
            ybar,
            height=.24\textwidth,
            width=.29\textwidth,
            grid style=dashed,
            bar width=1.1em,
            tick align=inside,
            ylabel style={yshift=-1em},
            symbolic x coords={{\knowledge,\reasoning}},
            enlarge x limits=0.55,
            xtick=data, 
            ymin=25, ymax=35,
            yticklabel style={/pgf/number format/precision=1, /pgf/number format/fixed zerofill}
        ]
            \addplot[fill=orange!40, draw] coordinates { ({\reasoning}, 25.93) ({\knowledge}, 34.84)};
            \addplot[fill=orange!85, draw] coordinates {({\reasoning}, 33.54) ({\knowledge}, 33.57)};
        \end{axis}
    }

\end{tikzpicture}